\documentclass[11pt,a4paper]{article}
\usepackage[a4paper,top=3cm,bottom=2cm,left=2cm,right=2cm,marginparwidth=1.75cm]{geometry}

\usepackage[english]{babel}
\usepackage[utf8x]{inputenc}
\usepackage[T1]{fontenc}
\usepackage{subcaption}
\usepackage{tikz}
\usepackage{amsmath}
\usepackage{amssymb}
\usepackage{graphicx}
\usepackage[colorinlistoftodos]{todonotes}
\usepackage[hyphens]{url}
\usepackage{hyperref}
\usepackage[hyphenbreaks]{breakurl}
\hypersetup{colorlinks=true,allcolors=blue,breaklinks=true}
\usepackage{listings}
\usepackage{xcolor}
\usepackage{csquotes}
\usepackage{pifont}% http://ctan.org/pkg/pifont

\newcommand{\cmark}{\ding{51}}%
\newcommand{\xmark}{\ding{55}}%

\definecolor{codegreen}{rgb}{0,0.6,0}
\definecolor{codegray}{rgb}{0.5,0.5,0.5}
\definecolor{codepurple}{rgb}{0.58,0,0.82}
\definecolor{backcolour}{rgb}{0.98,0.98,0.98}

\definecolor{turquoise}{cmyk}{0.65,0,0.1,0.1}
\definecolor{lilac}{cmyk}{0.0,0.19,0.0,0.22}

\usepackage[labelfont=bf]{caption}

\lstdefinestyle{mystyle}{
    backgroundcolor=\color{backcolour},   
    commentstyle=\color{codegreen},
    keywordstyle=\color{magenta},
    numberstyle=\tiny\color{codegray},
    stringstyle=\color{codepurple},
    basicstyle=\ttfamily\footnotesize,
    breakatwhitespace=false,         
    breaklines=true,                 
    captionpos=b,                    
    keepspaces=true,                 
    numbers=left,                    
    numbersep=5pt,                  
    showspaces=false,                
    showstringspaces=false,
    showtabs=false,                  
    tabsize=2
}

\newcommand{\R}{\mathbb{R}}
\newcommand{\mailtodomain}[1]{\href{mailto:#1@domain.com}{\nolinkurl{#1}}}

\title{
	\usefont{OT1}{bch}{b}{n}
	\huge On the Issues of TrueDepth Sensor Data for Computer Vision Tasks Across Different iPad Generations \\
}
\usepackage{authblk}
\author[1]{Steffen Urban$^*$}
\author[1]{Thomas Lindemeier$^*$}
\author[1]{David Dobbelstein}
\author[2]{Matthias Haenel}

\affil[1]{Carl Zeiss AG - Corporate Research Department}
\affil[2]{Onexip GmbH}
\affil[ ]{\{steffen.urban,thomas.lindemeier,david.dobbelstein\}@zeiss.com, haenel@onexip.com}

\begin{NoHyper}\footnotetext{These authors contributed equally to this work}
\end{NoHyper}

\begin{document}

\maketitle


%%%%%%%%%%%%%%%%%%%%%%%%%%%%%%%%%%%%%%%  
% Introduction
%%%%%%%%%%%%%%%%%%%%%%%%%%%%%%%%%%%%%%%  
\section{Introduction}
In 2017, Apple introduced the TrueDepth sensor with the iPhone X release \cite{face_id}. It's primary use case is biometric face recognition. The TrueDepth sensor module mainly consists of an infrared dot projector, an infrared camera and a RGB camera \cite{howtruedepthworks}. The depth is estimated by triangulating the projected dot pattern imaged by the infrared camera.

Although the main use case of the TrueDepth sensor module is biometric face recognition, the use of depth enriched image data for other computer vision tasks like segmentation \cite{scharwachter2013efficient}, portrait image generation \cite{portait_apple_doc}, 3D reconstruction and other metric measurements \cite{breitbarth2019measurement} is natural and lead to the development of various iOS apps \cite{hedges3d,theparallaxview,3dtruedepthcamerascan}.

In \cite{breitbarth2019measurement} the authors investigated the depth measurements accuracy of the TrueDepth sensor in an iPhoneX under a variety of different lighting and object. In \cite{vogt2021comparison} the authors compared the accuracy and usability of the TrueDepth and the Lidar sensor of an iPad Pro 2020 to an industrial 3D scanning system. In both publications the authors report measurement accuracy below a millimeter in a close measuring range. So far however, none of them report exactly how data was saved, pre- or post-processed and how if those results are repeatable across different devices and generations.

This report \textbf{reveals some significant inconsistencies in two different APIs on iPads} and makes the following contributions: 
\begin{itemize}
    \item A number of devices including iPhones from version 11 to version 13 and iPad 11'' 2gen to iPad 12.9'' 5gen is investigated (depicted in \ref{table:investigated_devices})
    \item We show two different ways to access the devices TrueDepth data. One using AVCapture session and one using an ARKit session and list various meta data for each device.
    \item We qualitatively and quantitatively examine the depth to RGB mapping and depth accuracy using a known object
    \item We identify two types of issues with the TrueDepth data access and it's reliability on all tested iPads. 
    \item We propose solutions to fix the data in a post-processing step.
    \item We make code to reproduce the results and a barebones iOS application to collect data available. \\
    Evaluation code: \url{https://github.com/ZEISS/iPad_TrueDepth_Issue_Eval}. \\ 
    iOS app: \url{https://github.com/ZEISS/iPad_TrueDepth_Issue_App}.
\end{itemize}

\begin{table*}[htb]
\centering
\begin{tabular}{|l|c|l|l|l|l|}
\hline
Device           & Nr. tested & iOS version & Chip & TrueDepth Camera  \\ \hline
iPad 11'' 2gen   &  2 & 14.x          & A12Z & 7 MP               \\ \hline
iPad 11'' 3gen   &  1 & 15.2.1      & M1   & 12 MP UWA          \\ \hline
iPad 12.9'' 4gen &  2 & 15.2.1    & A12Z & 7 MP               \\ \hline
iPad 12.9'' 5gen &  1 & 15.2.1        & M1   & 12 MP UWA          \\ \hline
 \hline 
iPhone 11 pro    &  2 & 15.2.1        & A13  & 12 MP              \\ \hline
iPhone 12        &  1 & 14.x             & A14  & 12 MP              \\ \hline
iPhone 12 pro    &  1 & 14.x             & A14  & 12 MP              \\ \hline
iPhone 13 pro    &  1 & 14.x           & A15  & 12 MP              \\ \hline
\end{tabular}
\caption{This table lists all tested iPads and iPhones. In some cases, we tested more than one device per generation.}
\label{table:investigated_devices}
\end{table*}

%%%%%%%%%%%%%%%%%%%%%%%%%%%%%%%%%%%%%%%  
% Preliminaries
%%%%%%%%%%%%%%%%%%%%%%%%%%%%%%%%%%%%%%%  
\section{Preliminaries}
This section recapitulates some computer vision basics that are needed to understand the investigations.
\subsection{Camera Intrinsics}
The camera matrix is a 3x3 matrix that maps a point $\textbf{p}=[x/z,y/z,1.0]^T$ from the camera coordinate system to a point $\textbf{u}=[u,v,1]^T$ in the image plane $\textbf{u}=\textbf{K}\textbf{p}$
Here $\textbf{K}$ is defined as:
\begin{equation}
    \textbf{K} = 
    \begin{bmatrix}
    f & s & c_x \\
    0.0 & f \cdot a & c_y \\
    0.0 & 0.0 & 1.0
    \end{bmatrix}
\end{equation}
and contains the focal length f in pixels, an aspect ratio $a$, as well as a shear factor $s$ and the principal point $c_x, c_y$. 

\begin{displayquote}
\textbf{In all devices that were tested, the factory calibrated intrinsics returned: $a=1$ and $s=0$. In addition, the principal point always coincides with the lens distortion center that can also be requested from the APIs \footnote{\url{https://developer.apple.com/documentation/avfoundation/avcameracalibrationdata/2881131-lensdistortioncenter/}}. Also compare listed meta data in table \ref{table:params_avsession} and table \ref{table:params_arsession}}
\end{displayquote}
\subsection{Lens distortion}

\begin{figure*}[ht]
  \centering
  \begin{subfigure}[t]{0.35\textwidth}
    \includegraphics[width=\textwidth]{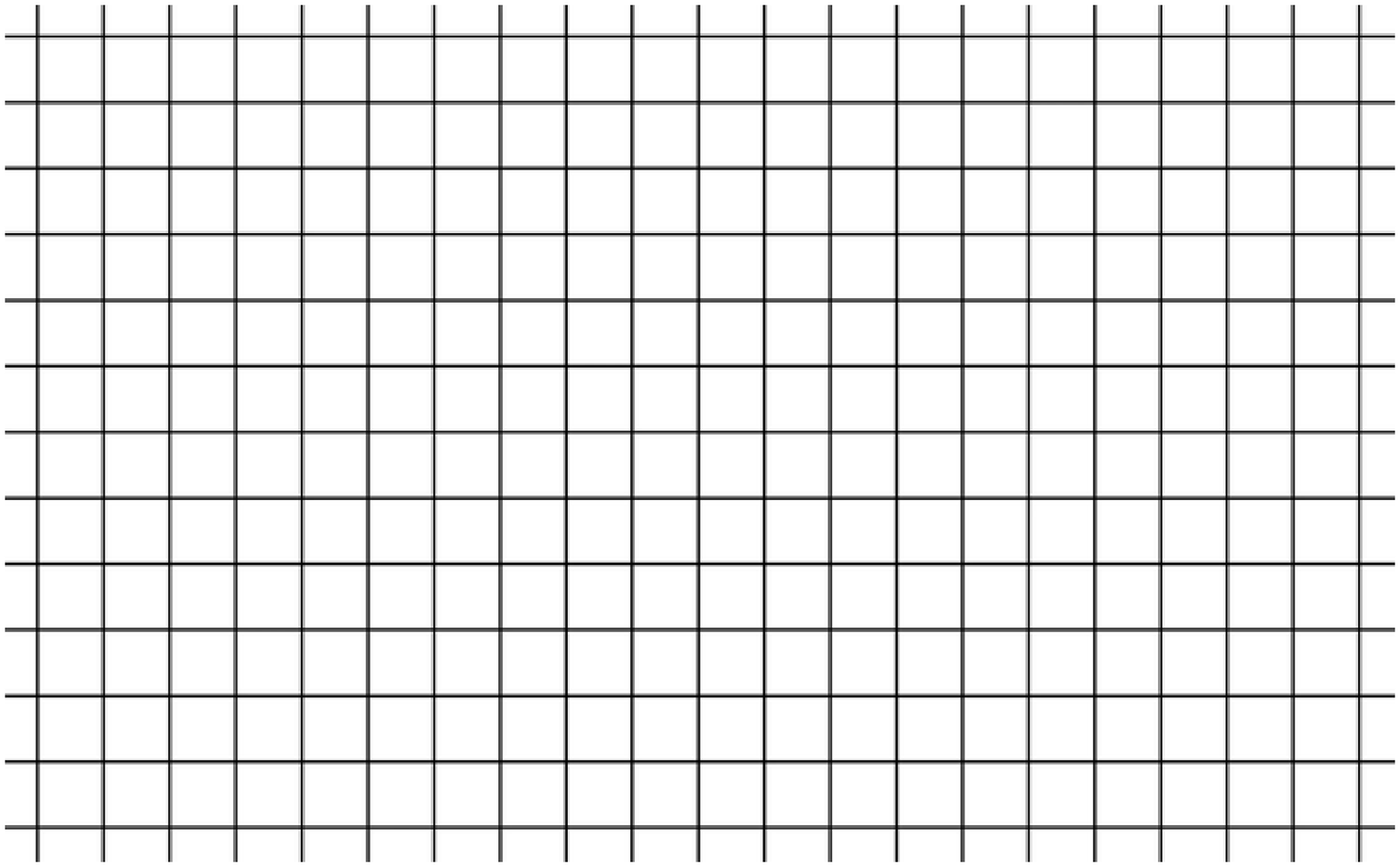}
    \caption{}
 \end{subfigure}
 \hspace{3cm}
  \begin{subfigure}[t]{0.35\textwidth}
    \includegraphics[width=\textwidth]{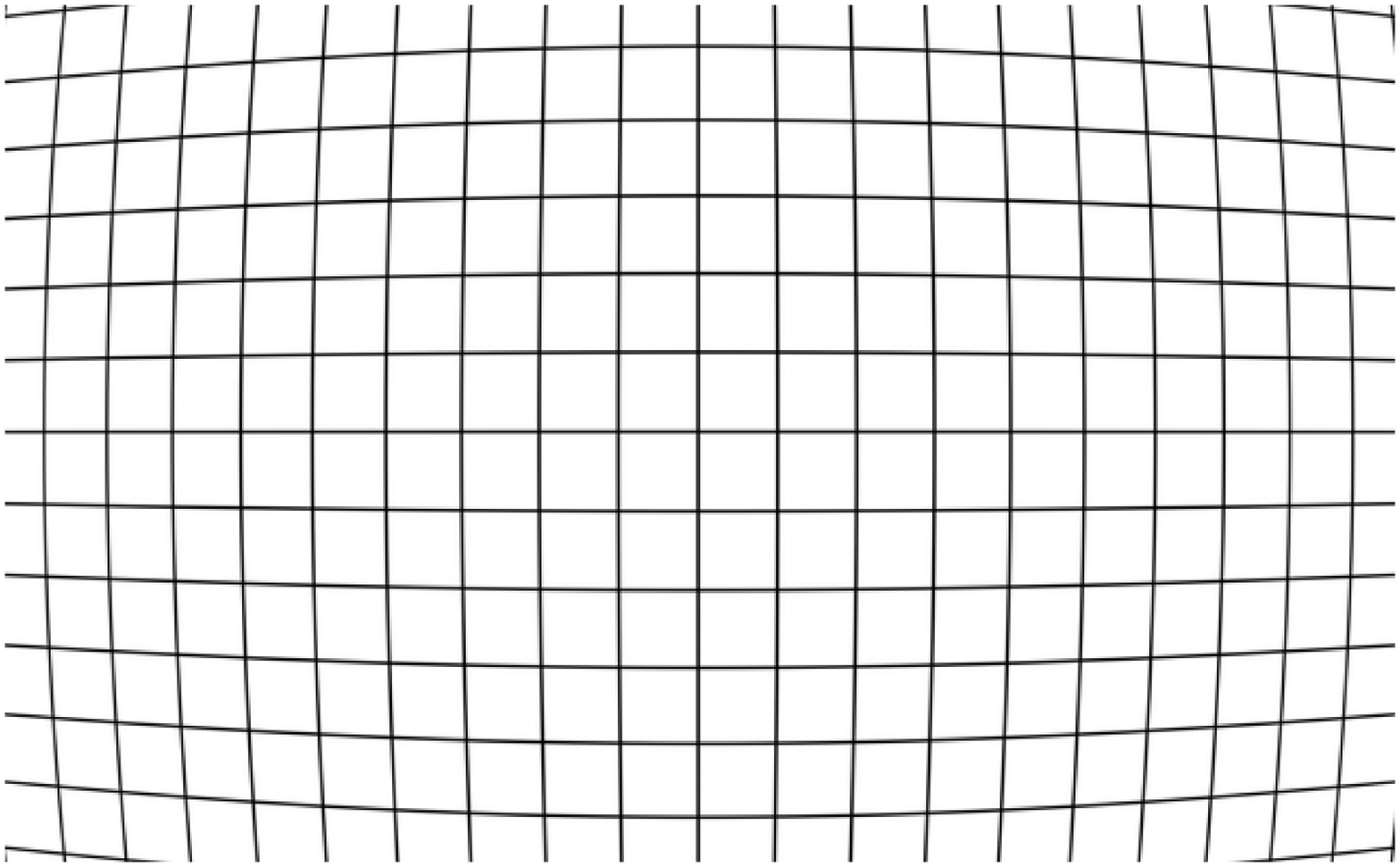}
    \caption{}
  \end{subfigure}
\caption{Example of radial distortions. The input grid (a) is radially distorted (b) and the lines becomes curves to bend outwards from the center (barrel distortion).}
\label{fig:distortion_illustration_comparison}
\end{figure*}

\textbf{Radial distortion:} This effect is a result of unequal refraction of light over the lens. Light rays at the edge of the lens are bent more than rays at the center of the lens. Straight lines become curves that either bent outwards (barrel distortion) or inwards (pincushion) from the center (see figure \ref{fig:distortion_illustration_comparison} (b)).\\

\textbf{Tangential distortion:} This is a result of sensors that are not parallel to the lens. The resulting images appear stretched and tilted. \\

The distortions can be removed by applying a transformation to the resulting images that warp the image data according to some distortion coefficients. The coefficients can be estimated with camera calibration. We evaluate the effect of the API provided distortion maps in \ref{sec:distortion}.

\subsection{Depth Image To Point Cloud}
To convert a depth image $D \in \R^{2}$ to a point cloud $\textbf{X}_{i,j} \in \R^3$ the following mapping can be used:
\begin{equation}
    \textbf{X}_{i,j} = D(i,j) \textbf{K}^{-1} \textbf{u}_{i,j}
\label{equation:unprojection}
\end{equation}
where $i=1..W$ and $j=1..H$ and $\textbf{u}_{i,j}=[i,j,1.0]^T$ 

A general observation is that:
\begin{displayquote}
\textbf{In all our tested devices the depth image has a resolution of 640x480 pixels, i.e. width $W=640$ and height $H=480$}.
\end{displayquote}

\subsection{Depth Camera to RGB Camera Transformation}
Usually when dealing with RGB-D data, not only the intrinsics of each camera need to be estimated but also the rotation and translation (extrinsics) that map the depth sensor frame to the RGB sensor frame \cite{basso2018robust}. This transformation is subsequently used to re-sample the depth image to the RGB image to obtain a per pixel depth measurement. For the TrueDepth sensor there is no way to access this extrinsic calibration information and we have to assume, that the transformation is accurately calibrated.

%%%%%%%%%%%%%%%%%%%%%%%%%%%%%%%%%%%%%%%  
% Accessing depth camera
%%%%%%%%%%%%%%%%%%%%%%%%%%%%%%%%%%%%%%%  
\section{Accessing The TrueDepth Data}\label{sec:true_depth}

We are aware of two different ways to access RGB-D data from the TrueDepth sensor and the RGB camera.
There is either the option to use the \textit{AVFoundation} \cite{avfoundation_framework} or the \textit{ARKit} framework \cite{arkit_framework}. \textit{AVFoundation} can be seen as a more low level interface, while \textit{ARKit} offers more abstractions and additional functionality relevant for AR applications. We found out that both frameworks output different meta data and images depending on the type of device that is used as can be seen in tables \ref{table:params_arsession} and \ref{table:params_avsession}.

The respective code that we used to retrieve meta and image data with both frameworks are listed in listing \ref{code:arkit} and listing \ref{code:av}. The code to process the metadata is shown in listing \ref{code:processmeta}. While \textit{ARKit} offers direct access to color intrinsics (listing \ref{code:arkit} line 38), the \textit{AVFoundation} provides this information only via further API calls (Listing \ref{code:av} lines 25-31).

\subsection{\textit{Retrieving Metadata}}
\begin{lstlisting}[language=swift]
func process(depthData: AVCameraCalibrationData, colorIntrinsicsMatrix: simd_float3x3, colorBuffer: CVPixelBuffer) { 
    let depthCalibrationData =  depthData.cameraCalibrationData
    
    let depthLensDistortionLookupTable = depthCalibrationData.lensDistortionLookupTable

    // Lens distortion center LDC [px]
    let depthLensDistortionCenter = depthCalibrationData.lensDistortionCenter
    // Intrinsics reference dimensions [px]
    let depthReferenceDimensions = depthCalibrationData.intrinsicMatrixReferenceDimensions
    // Deph intrinsics unscaled [px]
    let depthIntrinsicMatrix =  depthCalibrationData.intrinsicMatrix

    // scale the depth intrinsics
    var depthFormatDescription: CMFormatDescription?
                CMVideoFormatDescriptionCreateForImageBuffer(
                    allocator: kCFAllocatorDefault, 
                    imageBuffer: depthData.depthDataMap, 
                    formatDescriptionOut: &depthFormatDescription)
    let refX = Float(depthReferenceDimensions.width) / Float(depthFormatDescription!.dimensions.width)
    let refY = Float(depthReferenceDimensions.height) / Float(depthFormatDescription!.dimensions.height)
    let depthPrincipalPointX = depthIntrinsicMatrix.transpose.columns.0.z / refX
    let depthPrincipalPointY = depthIntrinsicMatrix.transpose.columns.1.z / refY
    let depthFocalLengthX = depthIntrinsicMatrix.transpose.columns.0.x / refX
    let depthFocalLengthY = depthIntrinsicMatrix.transpose.columns.1.y / refY

    // use these values in rgb-d 3d reconstruction
    self.processInternal(
        fx: depthFocalLengthX,  
        fy: depthFocalLengthY, 
        cx: depthPrincipalPointX, 
        cy: depthPrincipalPointY, 
        colorImage: colorBuffer, 
        depthMap:  depthData.converting(toDepthDataType: kCVPixelFormatType_DepthFloat32).depthDataMap)

}       

func processInternal(
        fx: Float,  
        fy: Float, 
        cx: Float, 
        cy: Float, 
        colorImage: CVPixelBuffer, 
        depthMap:  CVPixelBuffer) {
            // ...
}
\end{lstlisting}\label{code:processmeta}
\subsection{\textit{AVCaptureSession using AVFoundation}}\label{code:av}
\begin{lstlisting}[language=swift]
import AVFoundation

class RGBDViewController: AVCaptureDataOutputSynchronizerDelegate   {
    
    // ..

    func startAVSession() {
        self.session = AVCaptureSession()
        
        // some configurations
        // ...
        
        self.session.sessionPreset = AVCaptureSession.Preset.vga640x480
    }

    
     func dataOutputSynchronizer(_ synchronizer: AVCaptureDataOutputSynchronizer,
                                didOutput synchronizedDataCollection: AVCaptureSynchronizedDataCollection) {
        // some processing    
        // ...

        let depthData = syncedDepthData.depthData
        let sampleBuffer = syncedVideoData.sampleBuffer

        var colorMatrix = matrix_identity_float3x3
        guard let rgbPixelBuffer = CMSampleBufferGetImageBuffer(sampleBuffer) else  { return }
        if #available(iOS 11.0, *) {
            if let camData = CMGetAttachment(sampleBuffer, key: kCMSampleBufferAttachmentKey_CameraIntrinsicMatrix, attachmentModeOut:nil) as? Data {
                colorMatrix = camData.withUnsafeBytes { $0.pointee }
            }
        }
        
        process(depthData: depthData, colorIntrinsicsMatrix: colorMatrix, colorBuffer: sampleBuffer)
    }
}


\end{lstlisting}
\subsection{\textit{ARSession using ARKit}}\label{code:arkit}
\begin{lstlisting}[language=swift]
import ARKit

class RGBDViewController: ARSessionDelegate {

    // ...

    override func viewWillAppear(_ animated: Bool) {
        super.viewWillAppear(animated)
        
        startARKitSession()
    }
    
    func startARKitSession() {
        // create a session configuration
        let configuration = ARFaceTrackingConfiguration()
        // camera frame as reference
        configuration.worldAlignment = .camera
        for conf in ARFaceTrackingConfiguration.supportedVideoFormats {
            print(conf)
            if conf.imageResolution.height == 640 {
                configuration.videoFormat = conf
            }
        }
        print("using the following video format")
        print(configuration.videoFormat)

        self.sceneView.session.delegate = self

        // Run the view's session
        self.sceneView.session.run(configuration, options: [.resetTracking, .removeExistingAnchors] )
    }
    
    // MARK: - ARSessionDelegate
    
    func session(_ session: ARSession, didUpdate frame: ARFrame) {
        let depthData = frame.capturedDepthData
         // Color intrinsics [px]
        let colorIntrinsicsMatrix = frame.camera.intrinsics 
        
        if depthData != nil {
            process(depthData: depthData, colorIntrinsicsMatrix: colorIntrinsicsMatrix, colorBuffer: frame.capturedImage)
        }
    }
    
    // ...
}
\end{lstlisting}

%%%%%%%%%%%%%%%%%%%%%%%%%%%%%%%%%%%%%%%  
% Qualitative Overlap images
%%%%%%%%%%%%%%%%%%%%%%%%%%%%%%%%%%%%%%%  
\section{Evaluation of Depth Images and Depth to RGB Image Alignment}
Usually image processing tasks like segmentation, feature detection or edge detection are performed in the 2D RGB image. Some tasks like image registration, point cloud generation or foreground background subtraction can benefit from per pixel depth information. According to equation (\ref{equation:unprojection}) multiple factors influence the accuracy of the final 3D point \textbf{X}. Apart from a pixel accurate mapping (so that depth $D(i,j)$ actually corresponds to $i,j$ in the RGB image), the camera intrinsic matrix \textbf{K} influences the final 3D coordinate \textbf{X}.

\subsection{Qualitative Evaluation of Depth to RGB Alignment}
\begin{figure}
     \centering
     \begin{subfigure}[b]{0.33\textwidth}
         \centering
         \includegraphics[height=6.5cm]{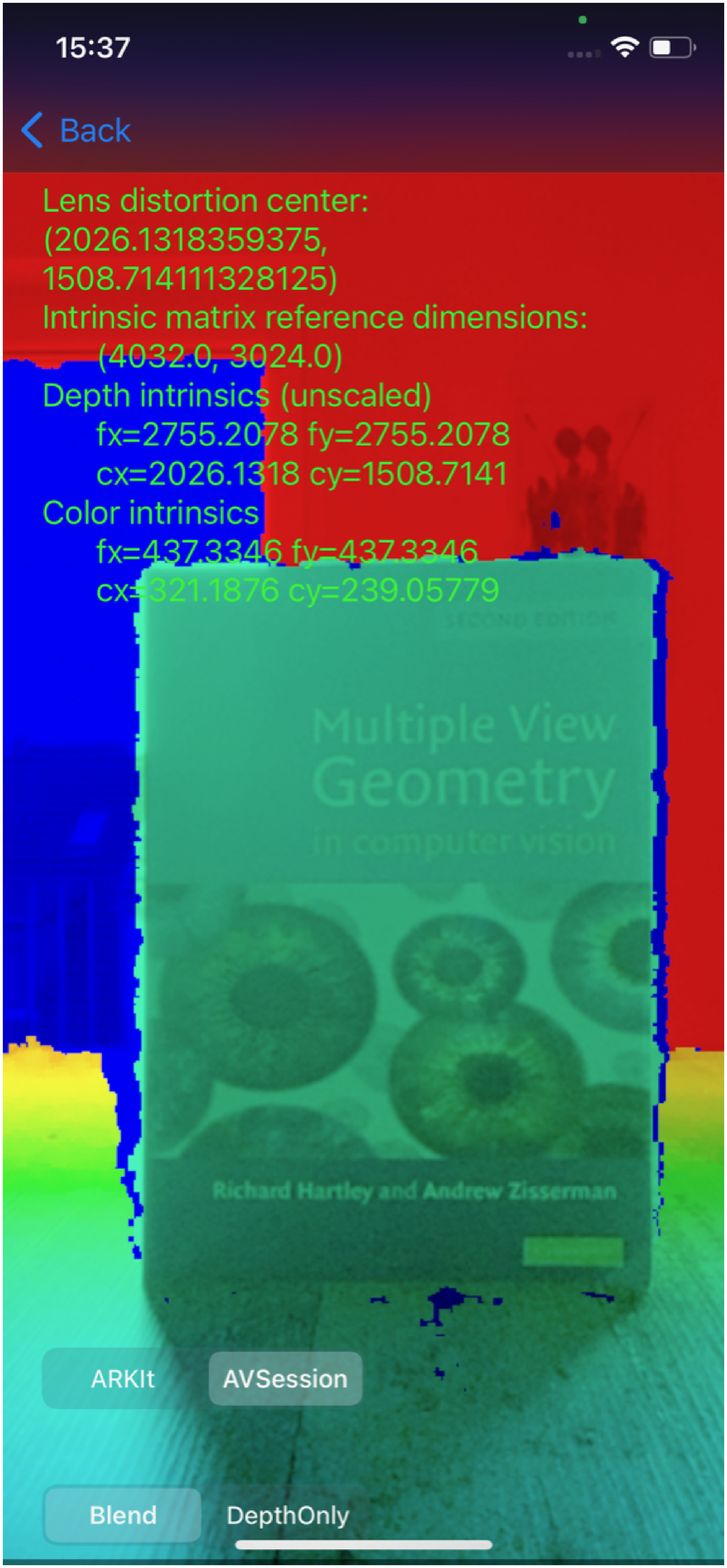}
         \caption{iPhone 11, AVSession}
         \label{fig:overlay_iphone11_avsession}
     \end{subfigure}
     %\hfill
     \begin{subfigure}[b]{0.33\textwidth}
         \centering
       \includegraphics[height=6.5cm]{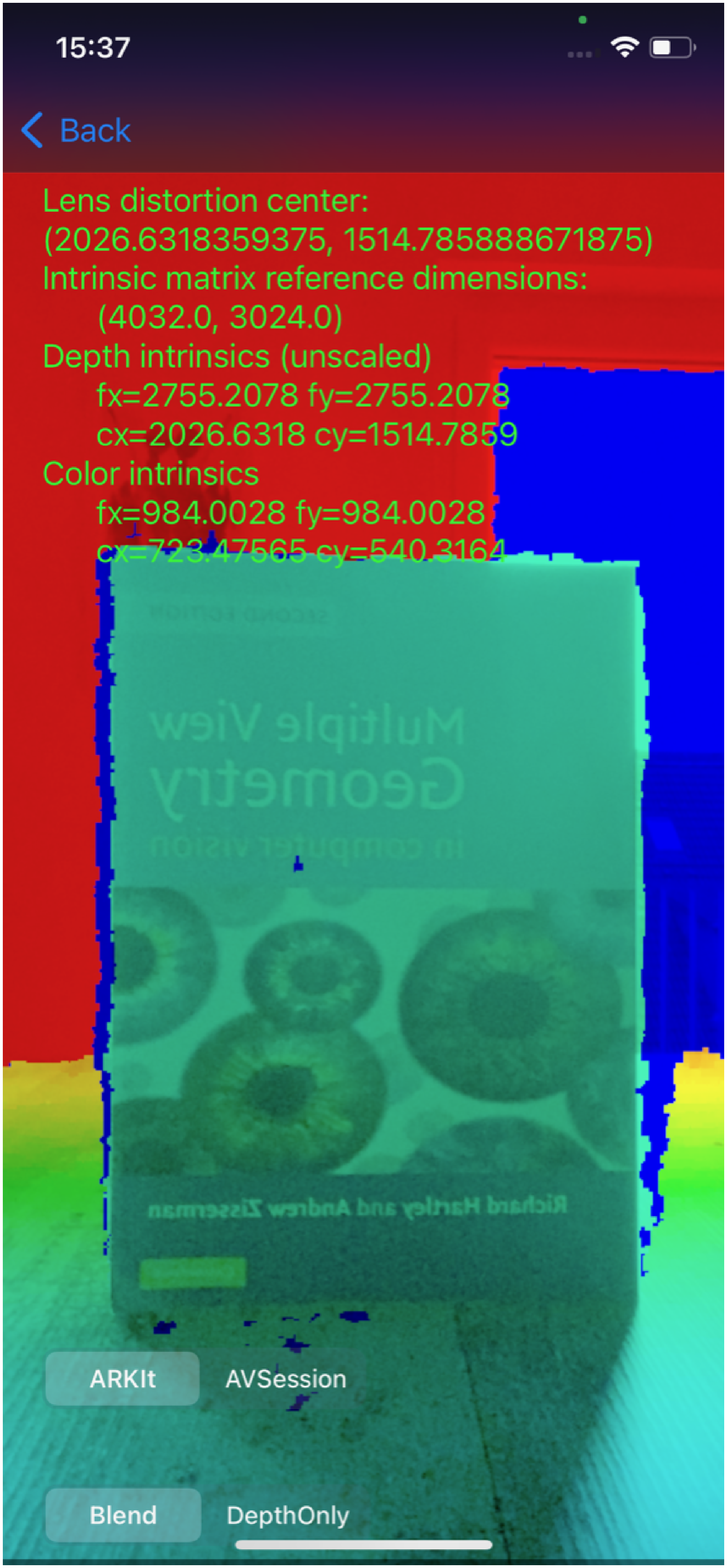}
         \caption{iPhone 11, ARKit}
         \label{fig:overlay_iphone11_arkit}
     \end{subfigure}\\
     %\hfill
     \begin{subfigure}[b]{0.33\textwidth}
         \centering
         \includegraphics[height=6.5cm]{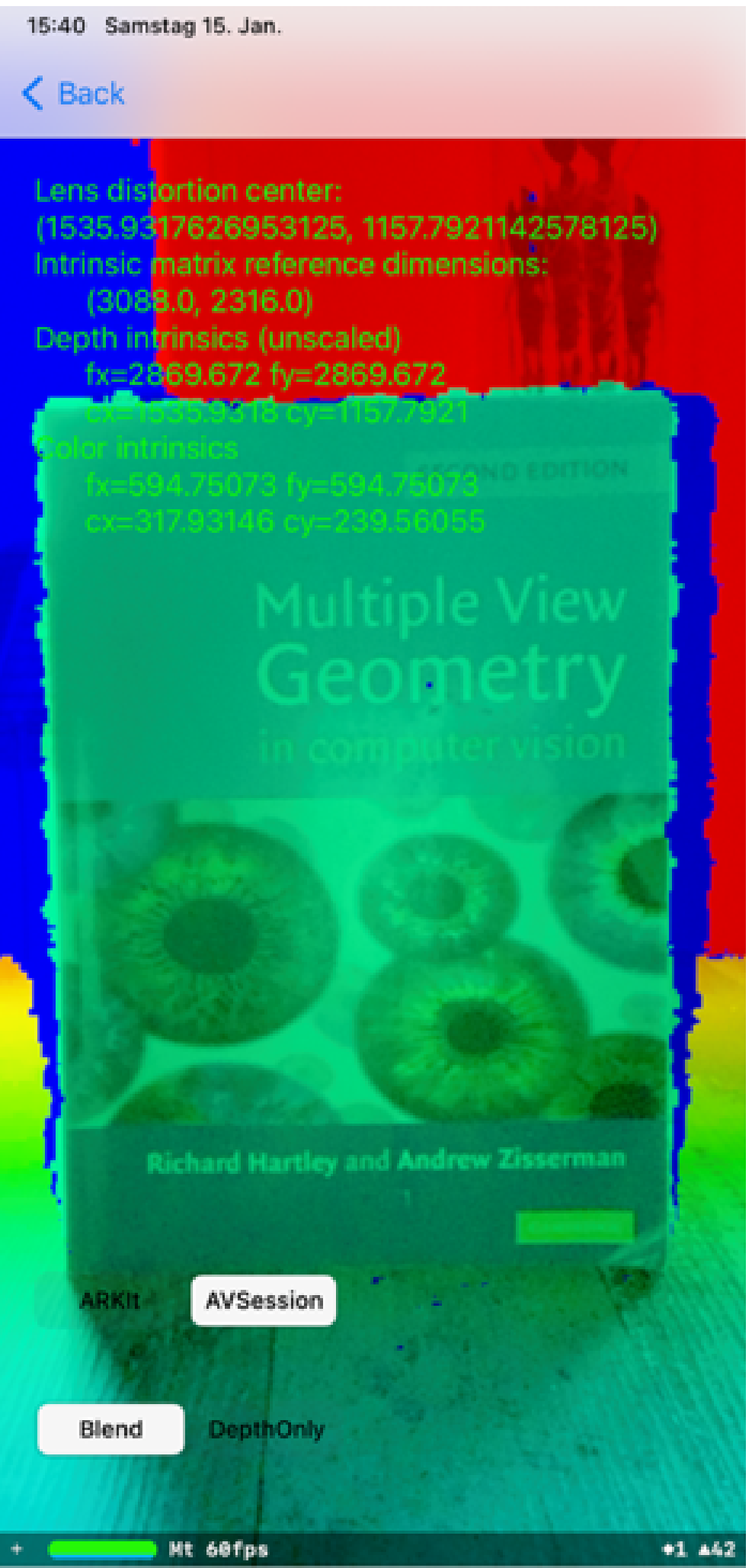}
         \caption{iPad 12.9'' 4gen, AVSession}
         \label{fig:overlay_iphone12_4gen_avsession}
     \end{subfigure}
     %\hfill
     \begin{subfigure}[b]{0.33\textwidth}
         \centering
       \includegraphics[height=6.5cm]{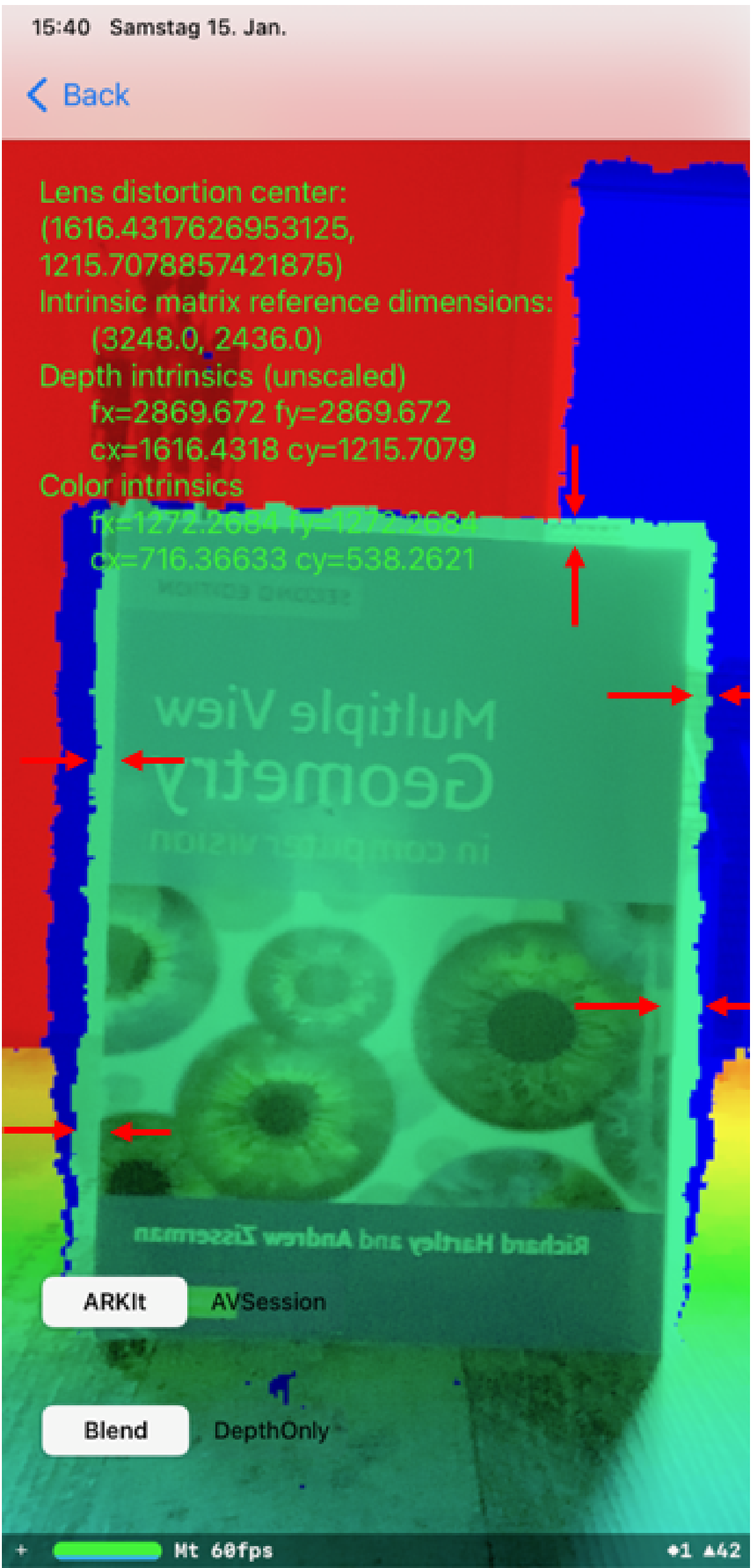}
         \caption{iPad 12.9'' 4gen, ARKit}
         \label{fig:overlay_iphone12_4gen_arkit}
     \end{subfigure} 
     %\hfill
     \begin{subfigure}[b]{0.33\textwidth}
         \centering
         \includegraphics[height=6.5cm]{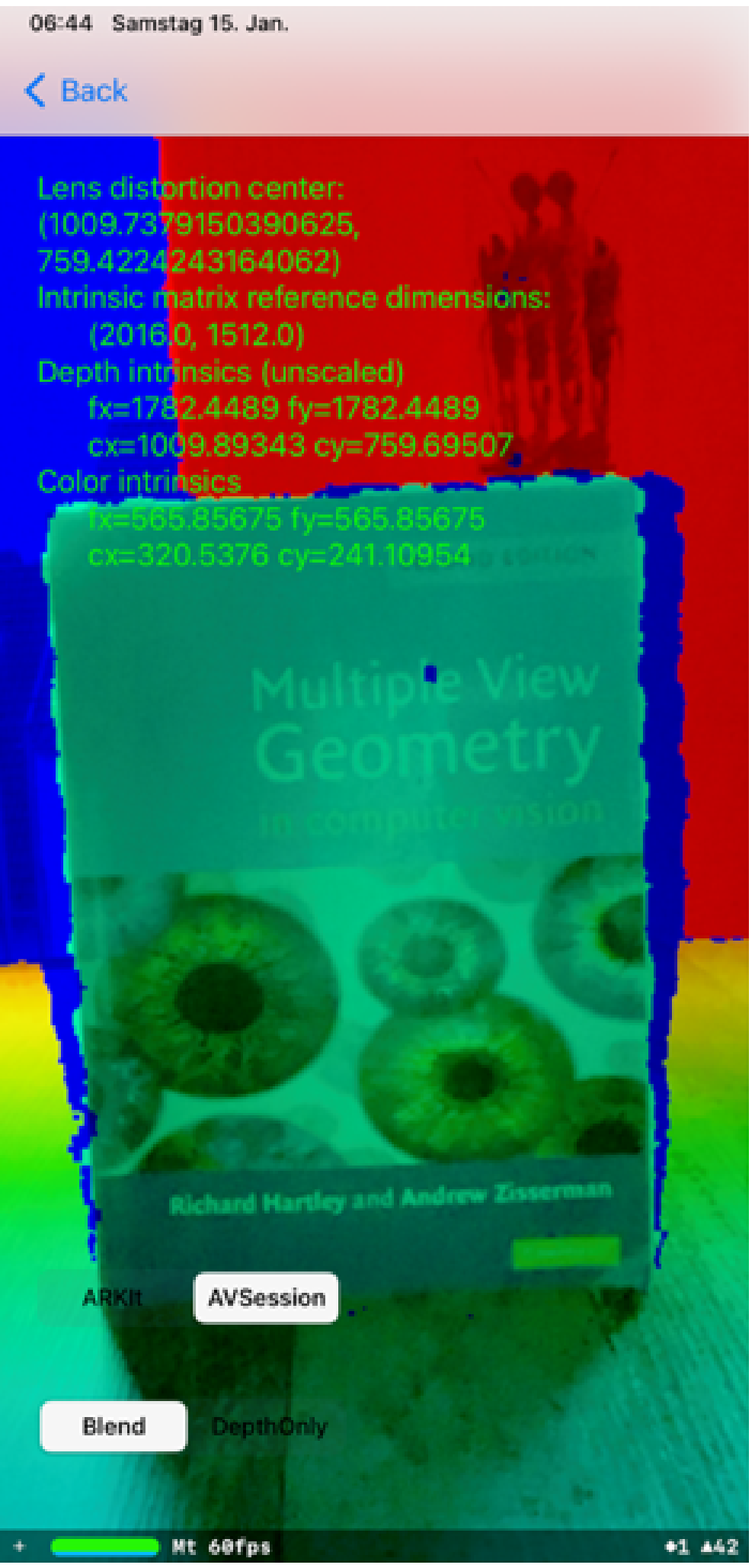}
         \caption{iPad 12.9'' 5gen, AVSession}
         \label{fig:overlay_iphone12_5gen_avsession}
     \end{subfigure}
     %\hfill
     \begin{subfigure}[b]{0.33\textwidth}
         \centering
         \includegraphics[height=6.5cm]{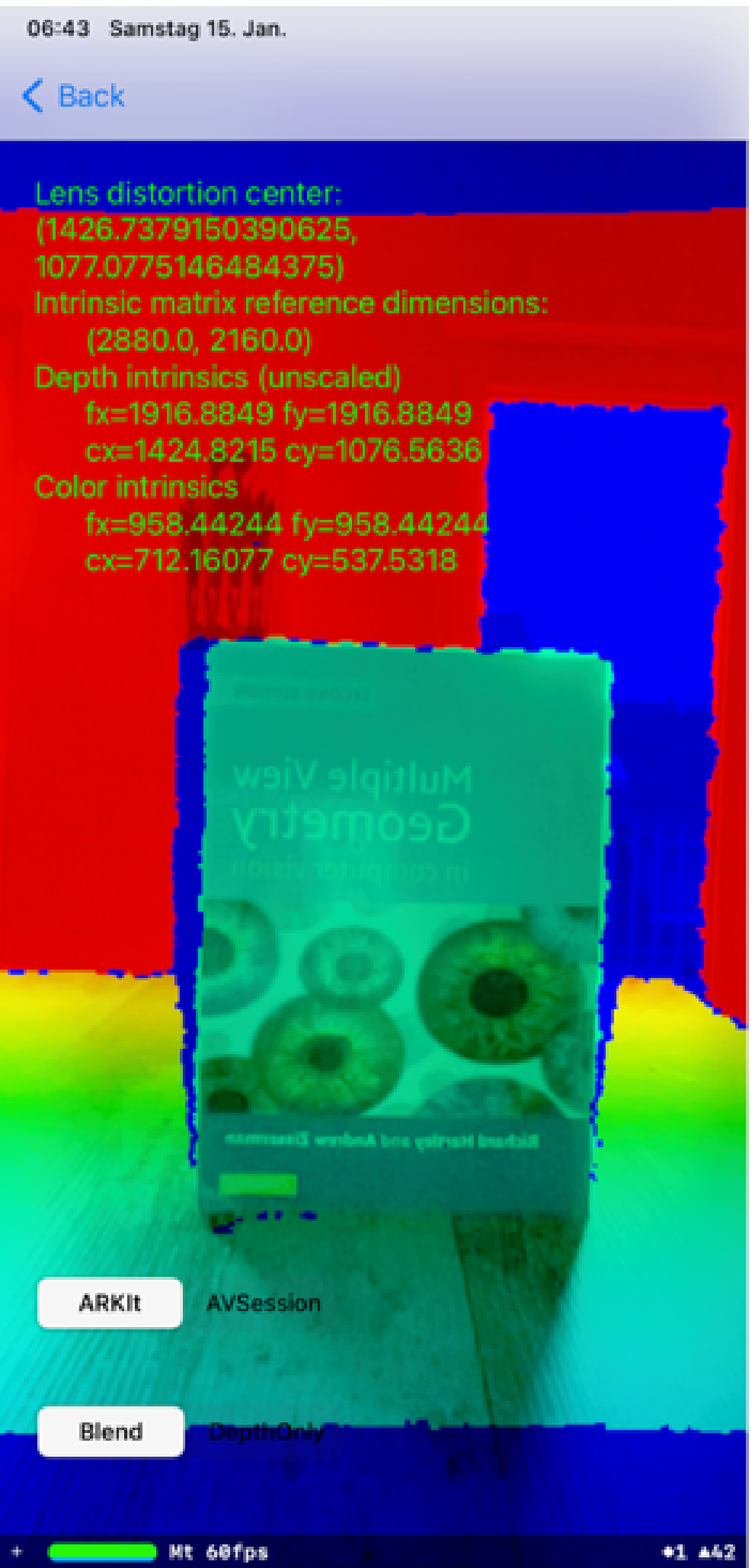}
         \caption{iPad 12.9'' 5gen, ARKit}
         \label{fig:overlay_iphone12_5gen_arkit}
     \end{subfigure}
     \caption{Qualitative evaluation of the depth to RGB mapping in AVSession (left column) and ARKit session (right column). The images were approximately taken from the same position. In most cases, the depth to RGB mapping seems to be well calibrated, expect for some noise along the border of the book. However the iPad12.9'' 4gen has issues during an ARKit session, depicted in figure (d). The overlay does not work correctly and the depth images appear zoomed in. For devices with ultra wide lenses like the iPad 12.9'' 5gen the field of view changes significantly in the ARKit session. Also in all cases the image is flipped around the up axis in ARKit sessions.}
     \label{fig:overlays}
\end{figure}

Without access to the raw depth images the extrinsic calibration can not be easily verified quantitatively. To perform a simple qualitative evaluation, we implemented an overlay application that superimposes the depth image color-coded on the RGB image. Figure \ref{fig:overlays} depicts overlays for both AVSession and ARKit session (left and right column) as well as for 3 different devices. \\
For iPhone 11 Pro the overlay looks fine in each mode apart from noise in the depth image along the edges of the book. However in all tested devices the iPad 11'' 2gen as well as the iPad 12.9'' 4gen have significant differences in the overlay if the data is recorded in ARKit mode as depicted in figure \ref{fig:overlay_iphone12_4gen_arkit}. This will cause accuracy issues, if multiple depth maps are fused from different viewpoints or are used for background segmentation. 

To subsume our qualitative observation: 
\begin{displayquote}
\textbf{The alignment looks reasonable in ALL tested devices in an AVSession. Only if the data is recorded in an ARKit session the alignment is not correct for iPad 11'' 2gen and iPad 12.9'' 4gen. The depth images are too wide. In section \ref{sec:solutions}, we propose a method to correct this issue using the provided API data.}
\end{displayquote}

\subsection{Accuracy of Unprojected Depth}
\begin{figure}
     \centering
      \includegraphics[width=\textwidth]{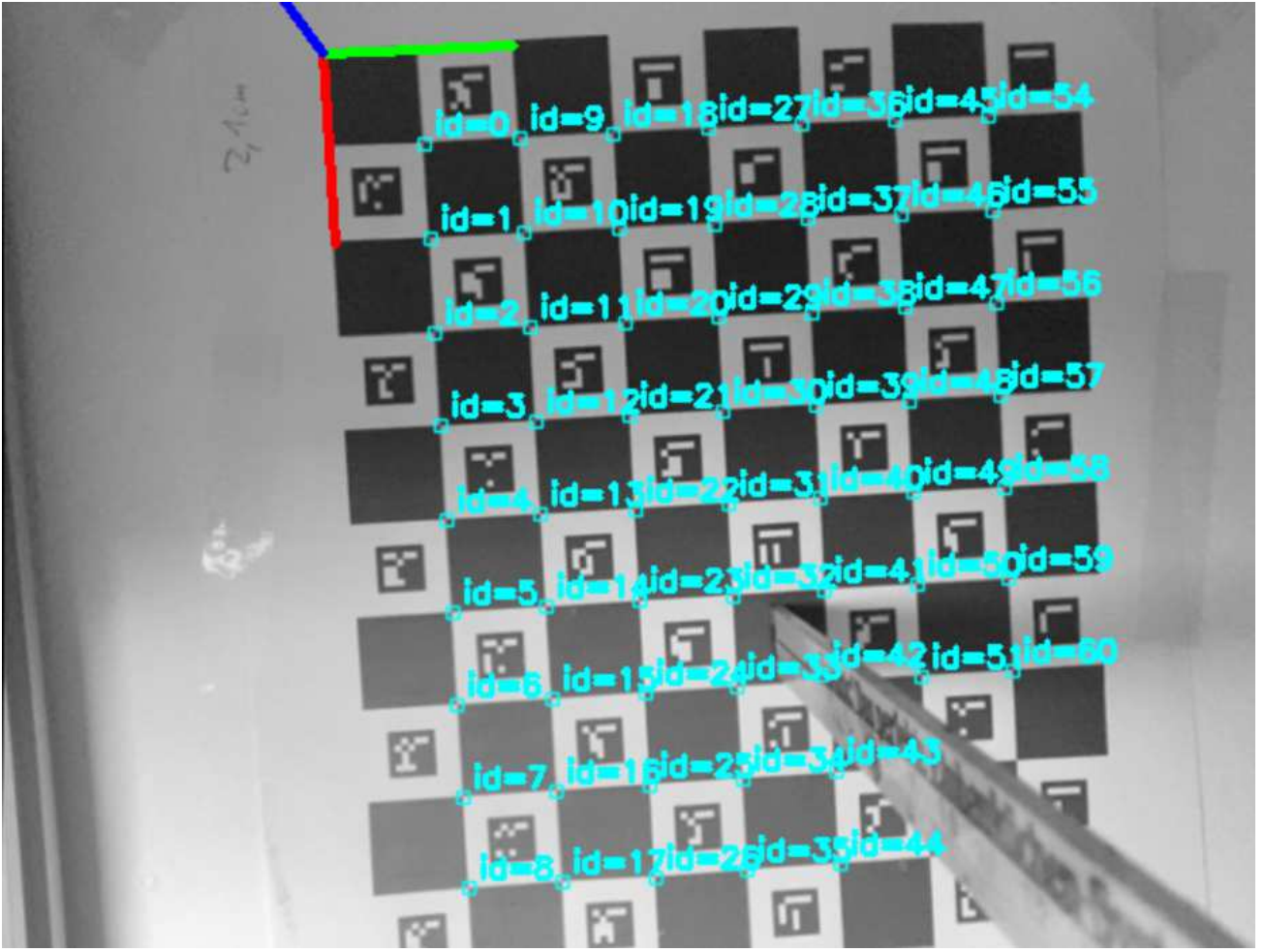}
      \caption{Depth verification setup. Images are taken from a metrically known Charuco board. Then Charuco corners $\textbf{u}_{charuco}$ are extracted in the RGB image and the camera pose $\textbf{T}^c_{charuco}$ is estimated.}
    \label{fig:frontal_image}
\end{figure}

After revealing issues with the Depth to RGB alignment for the iPad 11'' 2gen, we investigated the quality of the depth values for all devices. According to equation (\ref{equation:unprojection}) the z-component of the un-projected 3D point is not affected by wrong camera intrinsics. Thus we only compare the z-component $Z^c_{av}$ of the un-projected 3D point $\textbf{X}^c_{av}=[X^c_{av},Y^c_{av},Z^c_{av}]^T$ to the z-component $Z^c_{ch}$ of the 3D points from the known Charuco board in the camera coordinate system: $\textbf{X}^c_{ch}=[X^c_{ch},Y^c_{ch},Z^c_{ch}]^T$. Latter however is influences by camera intrinsics as it relies on camera pose estimation to rotate the board coordinates $\textbf{X}_{ch}$ to the camera coordinate system. Ideally the z-components of both 3D points should be close ($Z^c_{av}\approx Z^c_{ch}$ indicating that both depth values and factory intrinsics are reliable. We use the following procedure to record reference data (all data is recorded in AVSession):
\begin{enumerate}

\item Take frontal images of a Charuco checkerboard (see Figure \ref{fig:frontal_image}).

\item Extract Charuco corners $\textbf{u}_{ch}$ in the RGB image that correspond to the metrically known 3D Charuco board points $\textbf{X}_{ch}$.

\item Estimate camera pose $\textbf{T}^c_{ch}$ (using OpenCV's \textit{solvePnP} function with \textit{SOLVEPNP\_ITERATIVE}) that transforms a 3D point from the Charuco board to the camera coordinate system yielding known 3D points in the camera coordinate system: $\textbf{X}^c_{ch}=\textbf{T}^{c}_{ch}\textbf{X}_{ch}$. The image points are normalized using either factory intrinsics or our own Charuco calibrated intrinsics.

\item Calculate 3D points $\textbf{X}^c_{av}$ by unprojecting the 2D Charuco corners $\textbf{u}_{ch}$ to the camera coordinate system using equation (\ref{equation:unprojection}) and bilinear interpolation of depth values.

\end{enumerate}

\begin{figure}
     \centering
     \begin{subfigure}[b]{\textwidth}
         \centering
         \includegraphics[width=0.35\textwidth]{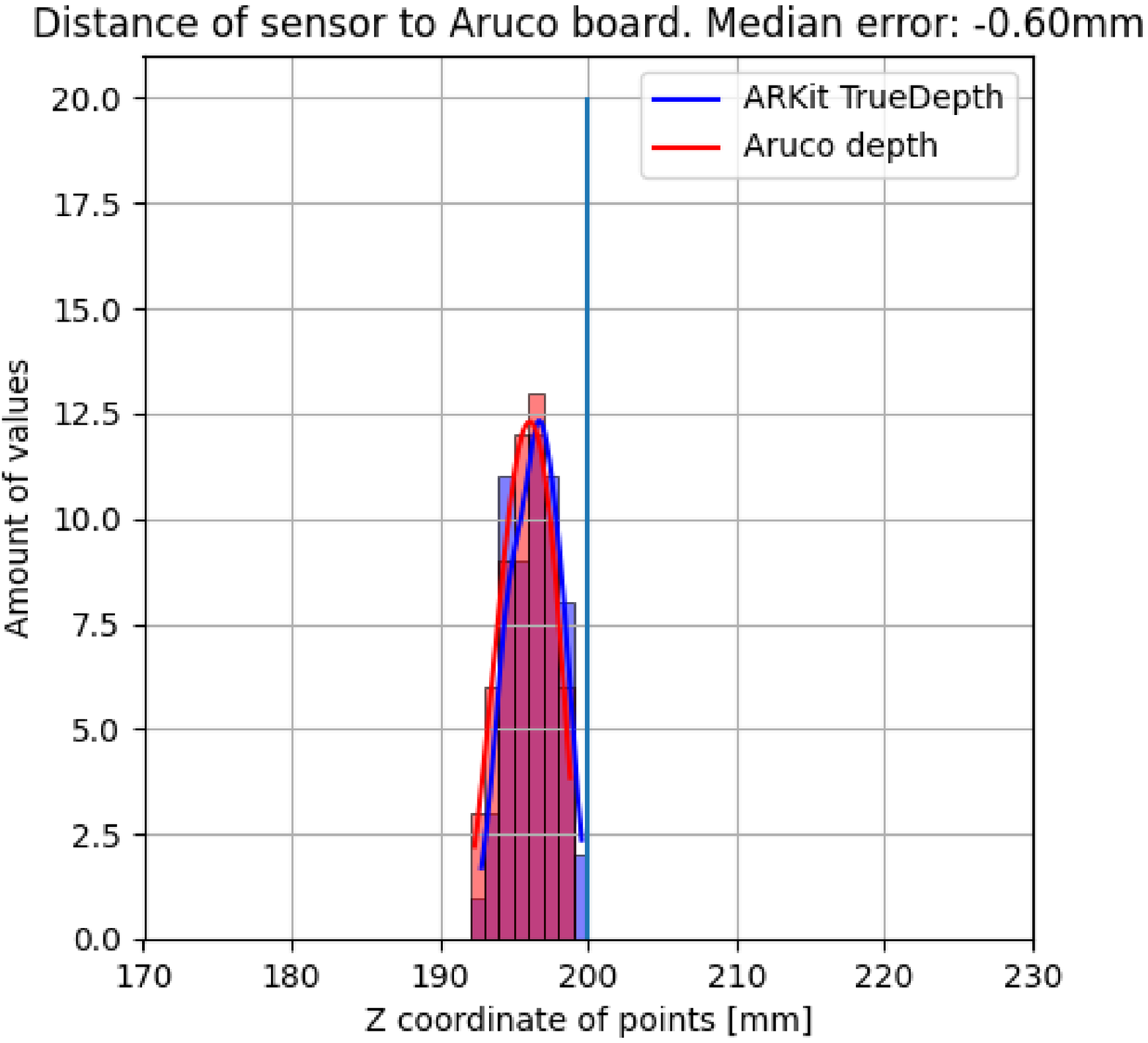}\quad
         \includegraphics[width=0.35\textwidth]{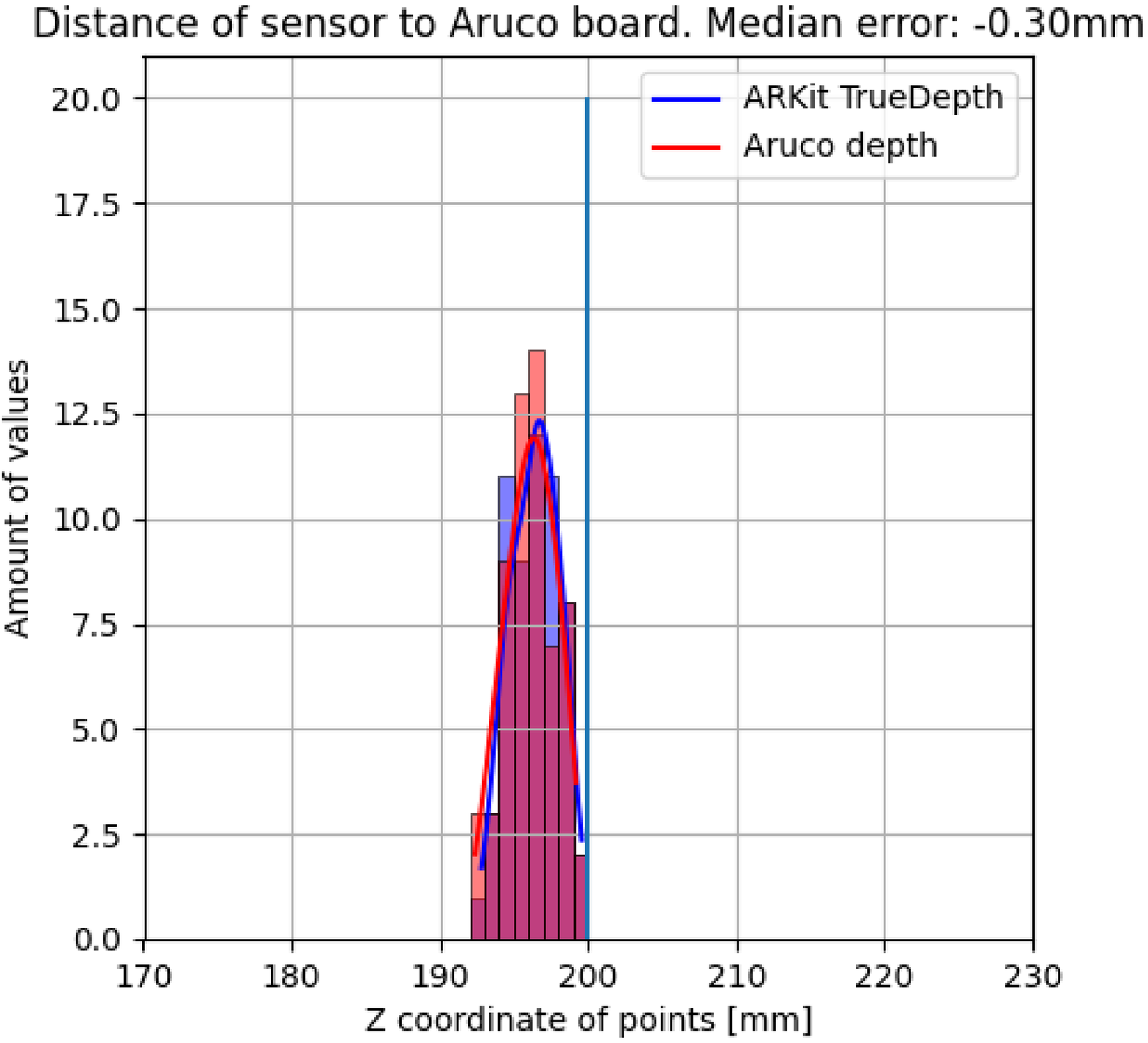}
         \caption{\textbf{iPhone 11 Pro}. Left: AVSession factory intrinsics. Right: own Charuco intrinsics.}
         \label{fig:depth_error_iphon11}
     \end{subfigure}
     \hfill
     \begin{subfigure}[b]{\textwidth}
       \centering
         \includegraphics[width=0.35\textwidth]{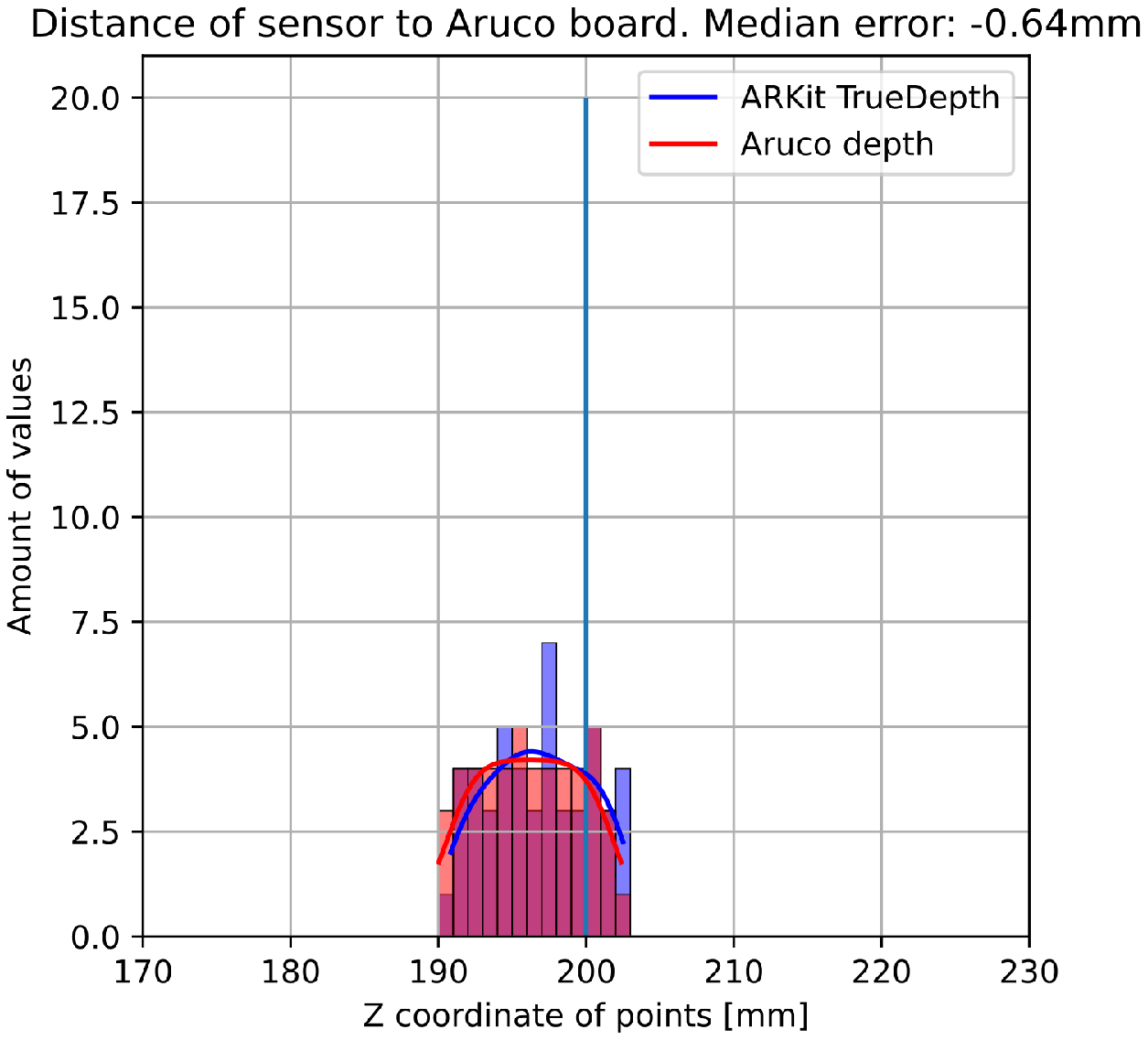}\quad
         \includegraphics[width=0.35\textwidth]{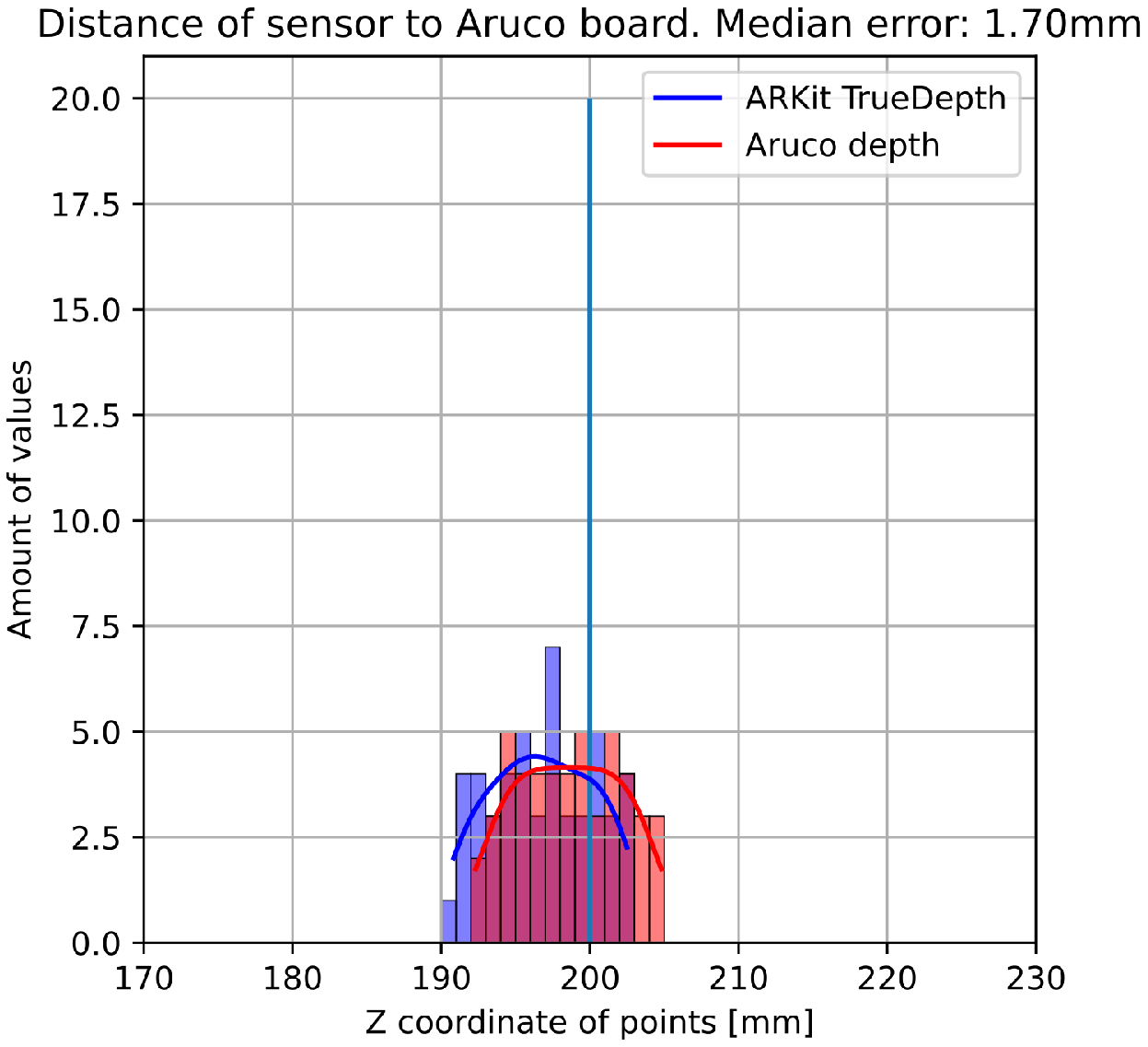}
         \caption{\textbf{iPad 12.9'' 4gen}. Left: AVSession factory intrinsics. Right: own Charuco intrinsics.}
         \label{fig:depth_error_ipad12_4gen}
     \end{subfigure}
     \hfill
     \begin{subfigure}[b]{\textwidth}
         \centering
         \includegraphics[width=0.35\textwidth]{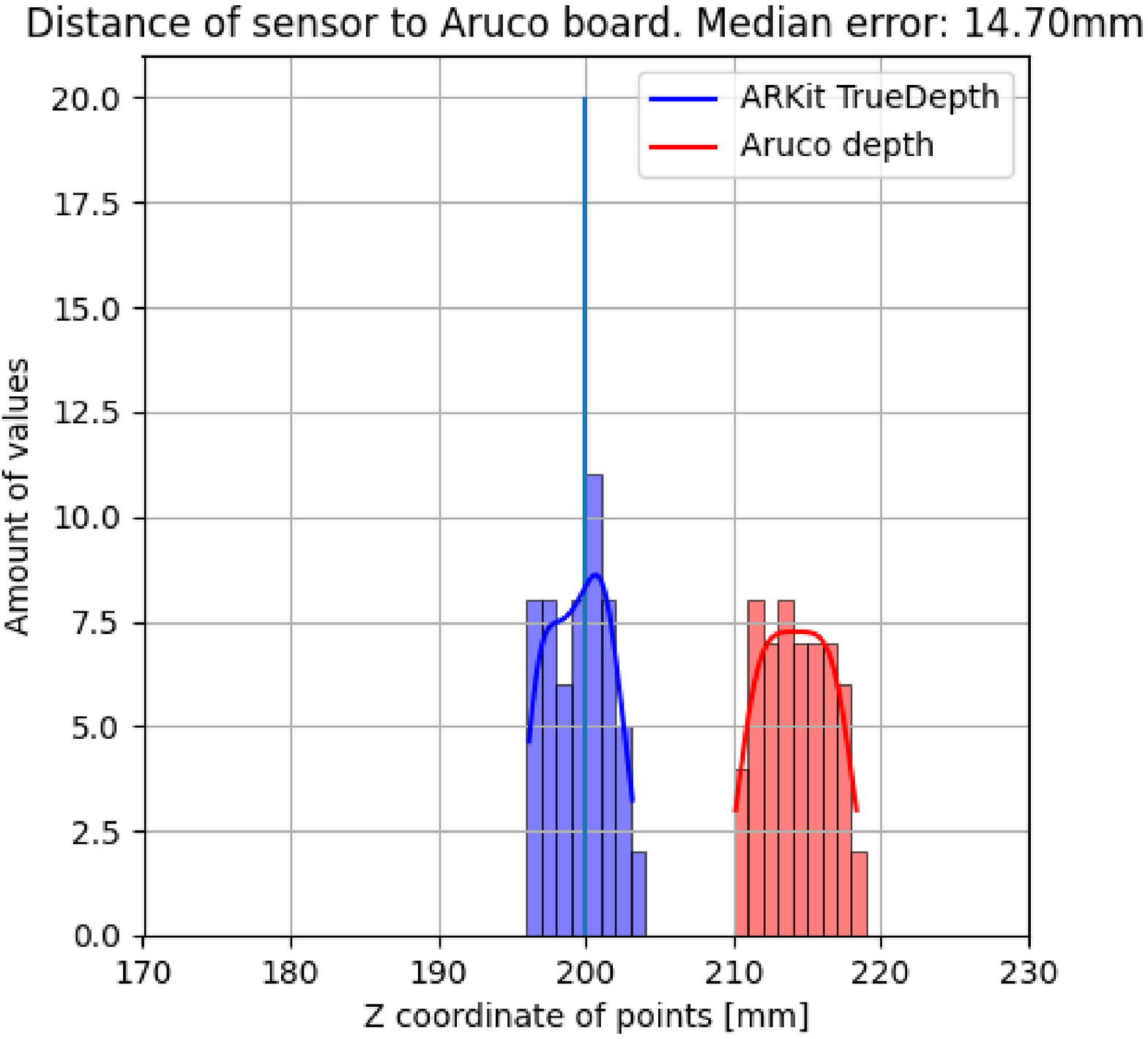}\quad
         \includegraphics[width=0.35\textwidth]{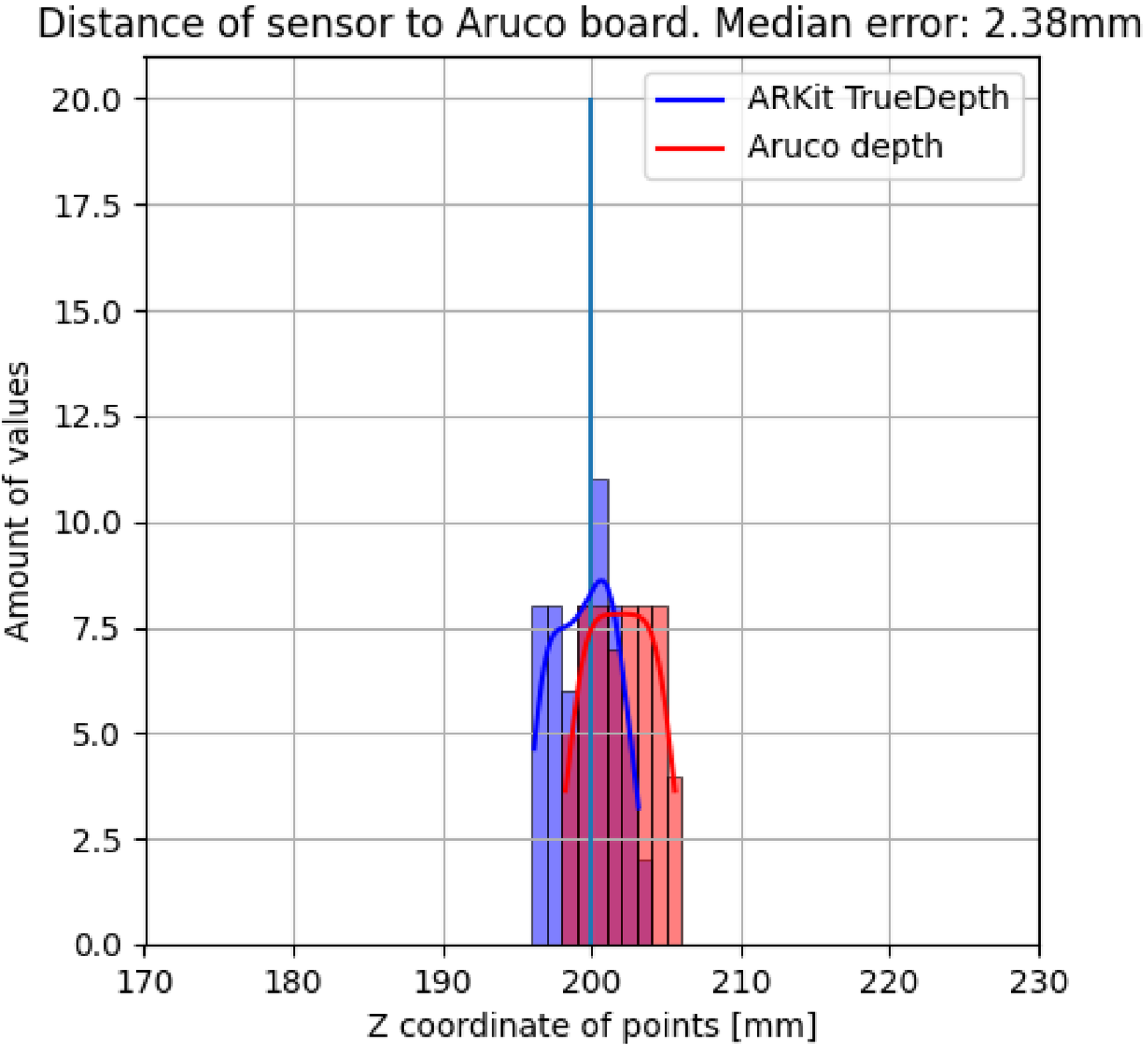}
         \caption{\textbf{iPad 12.9'' 5gen}. Left: AVSession factory intrinsics. Right: own Charuco intrinsics.}
         \label{fig:depth_error_ipad12_5gen}
     \end{subfigure}

     \caption{Quantitative evaluation of intrinsics and depth measurement. The data was recorded in an \textbf{AVSession}. With the correct intrinsics the two distributions should align as closely as possible. The devices were held in an approximate distance of 200mm to the Charuco board. Depicted are histograms of the Z component of ARKit $\textbf{X}^c_{arkit}$ and Charuco $\textbf{X}^c_{charuco}$ points, i.e. the depth to the board. The left column uses the calibration given by the AVSession API. The right column uses our own Charuco calibrated values. For the iPhone 11 Pro (figure (a)) and the iPad 12.9'' 4gen (figure (b)) it is clearly visible that both align very well. The iPad 12.9'' 5gen shows significant disagreement if the AVSession factory intrinsics are used to estimate the camera pose as depicted in figure (c) left. The raw depth measurements, however, seem to be correct.}
     \label{fig:depth_error}
\end{figure}

Under perfect intrinsics and depth values: $\textbf{X}^c_{av}=\textbf{X}^c_{ch}$. However, due to uncertainties and noise in corner detection, depth and camera pose estimation: $\textbf{X}^c_{av}\approx\textbf{X}^c_{ch}$. So in figure \ref{fig:depth_error}, we report the difference $d=Z^c_{av}$-$Z^c_{ch}$ in millimeters.

To subsume our quantitative observation: 
\begin{displayquote}
\textbf{For ALL iPhones (depicted is only iPhone 11 Pro) and iPad 12.9'' 4gen and iPad 11'' 2gen (depicted is only iPad 12.9'' 4gen), $Z^c_{av}$ and $Z^c_{ch}$ agree within a millimeter (or even less) if the factory intrinsics are used. For iPad 11'' 3gen and iPad 12.9'' 5gen (only iPad 12.9'' 5gen is depicted) there is a large difference (>14mm) if the factory intrinsics are used to estimate the camera pose. Using our own intrinsic Charuco calibration the problem vanishes as depicted in figure \ref{fig:depth_error_ipad12_5gen}. \textit{This indicates wrong camera intrinsics but correct depth values.}}
\end{displayquote}

%%%%%%%%%%%%%%%%%%%%%%%%%%%%%%%%%%%%%%%  
% Calibration Section
%%%%%%%%%%%%%%%%%%%%%%%%%%%%%%%%%%%%%%%  
\section{Verifying Intrinsic Camera Calibration}
In the last chapter, we pointed out that there is a problem with the factory calibrated intrinsics for 2 types of iPad devices (iPad11'' 3gen and iPad 12.9'' 5 gen). To verify the factory calibration of the RGB camera that can be queried from the AVSession API, we perform two test:
\begin{enumerate}
    \item We perform standard camera calibration using an Charuco checkerboard \cite{garrido2016aruco} to evaluate the factory calibrated focal length.
    \item We apply the forward and inverse distortion lookup tables to all devices and qualitatively check the results to verify that the images are already undistorted.
\end{enumerate}
The code to reproduce the results from this chapter can be found online: \\ \url{https://github.com/ZEISS/iPad_TrueDepth_Issue_Eval}.
\subsection{Verifying Factory Focal Length}
To check the factory calibrated intrinsic values, we perform standard camera calibration using OpenCV's \textit{calibrateCameraCharuco} function. For each device, we record datasets with $>50$ images while  moving the device around a Charuco board. To reduce rolling shutter artifacts, we set a acquisition delay of 500ms and slowly move the device around the Charuco board. To ensure spatial variability and avoid clusters of camera positions which would make camera calibration unstable, we use a simple voxel grid approach to reject images that have been taken from a very similar viewpoint. The pose of each image is estimated using OpenCV's \textit{solvePnP} function with \textit{SOLVEPNP\_ITERATIVE} flag and the image points are normalized using the intrinsics provided by the API. In addition, we reject an image for calibration if less then 10 Charuco corners have been detected. 

For each device, we only calibrate the focal length, i.e. we fix aspect ratio, distortion and principal point optimization. In addition, we use the AVSession API provided intrinsics as an initial guess. 

The results are depicted in table \ref{tab:cam_calib} and indicate:
\begin{displayquote}
\textbf{that the factory intrinsics, that we get from the AVSession API are off by 6-7\% for iPad 12.9'' 5gen and iPad11'' 4gen. For all other devices the calibrated focal length agrees well with the Charuco calibration and the difference is within 1\% on average.}
\end{displayquote}

Having a look at table \ref{table:params_avsession} and table \ref{table:params_arsession} the 7\% can be found again if the ratio between the unscaled depth intrinsics of both APIs is calculated (which is about 7.5\%). This issue only occurs for iPad 12.9'' 5gen and iPad11'' 4gen. Both iPads have a 12MP UWA TrueDepth camera.

Another observation is:
\begin{displayquote}
\textbf{that the update of the devices from iOS 14 to iOS 15 slightly changed the factory calibrated focal length parameters, also depicted in table \ref{table:params_avsession} and \ref{table:params_arsession}.}
\end{displayquote}
This is potentially problematic for applications relying on these factory calibrated camera parameters and performed verification steps based on them.

\subsection{Applying Inverse Distortion Lookup Table}
\label{sec:distortion}
Using the API it is possible to acquire forward and inverse distortion lookup tables. 
According to the documentation the depth data is always mapped to the same distortion as the RGB camera in order for both to align in pixel space\footnote{\url{https://developer.apple.com/documentation/avfoundation/avcameracalibrationdata/2881129-lensdistortionlookuptable}}: 

\begin{displayquote}
When dealing with \textit{AVDepthData} objects, the disparity/depth map representations are geometrically distorted to align with images produced by the camera.
\end{displayquote}

\begin{figure*}[t!]
  \centering
  \begin{subfigure}[t]{0.3\textwidth}
    \includegraphics[width=1.0\textwidth]{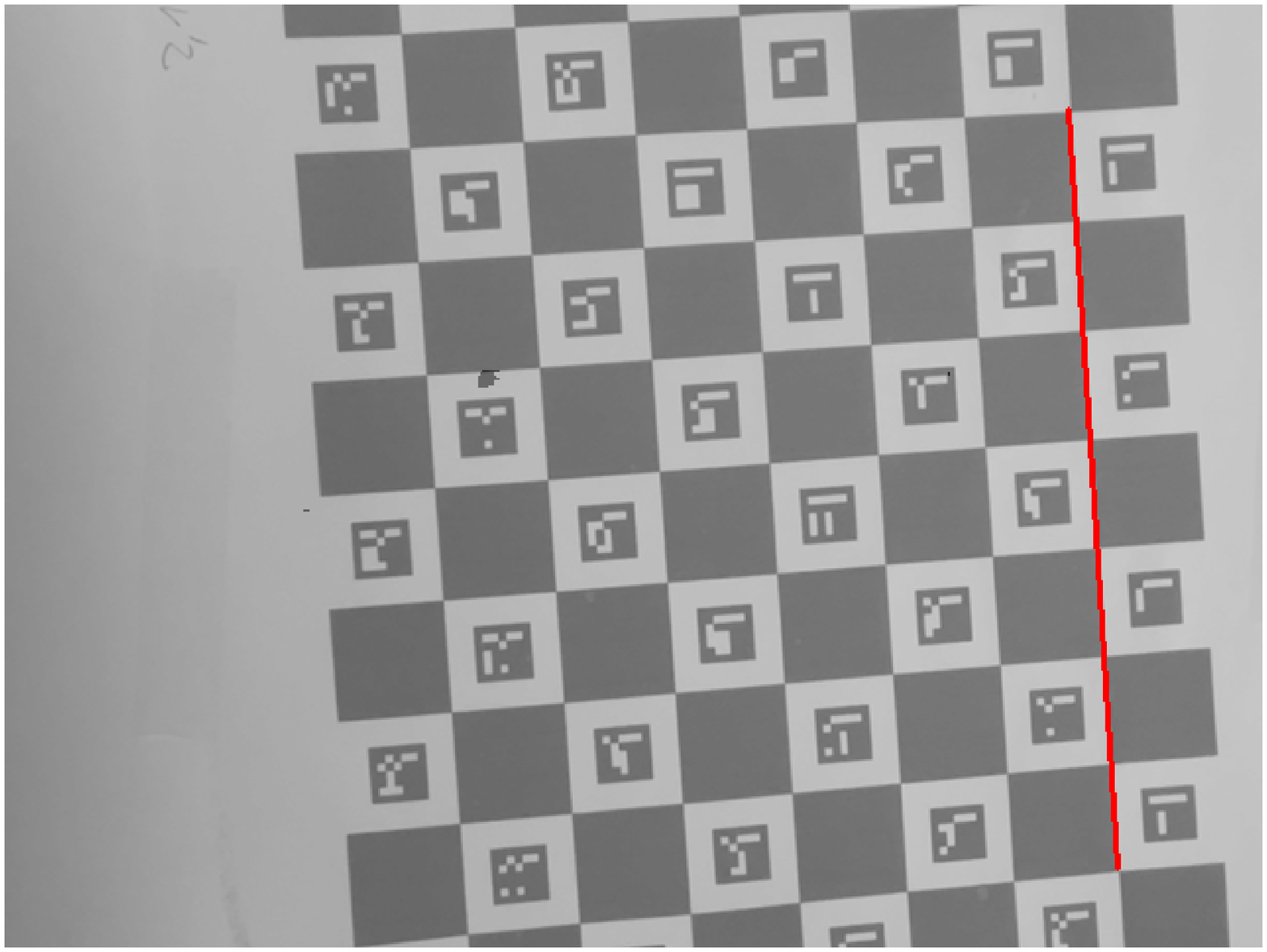}
    \caption{Original image acquired from AvCaptureSession}
    \label{fig:distortion_comparison_a}
 \end{subfigure}
  \begin{subfigure}[t]{0.3\textwidth}
    \includegraphics[width=1.0\textwidth]{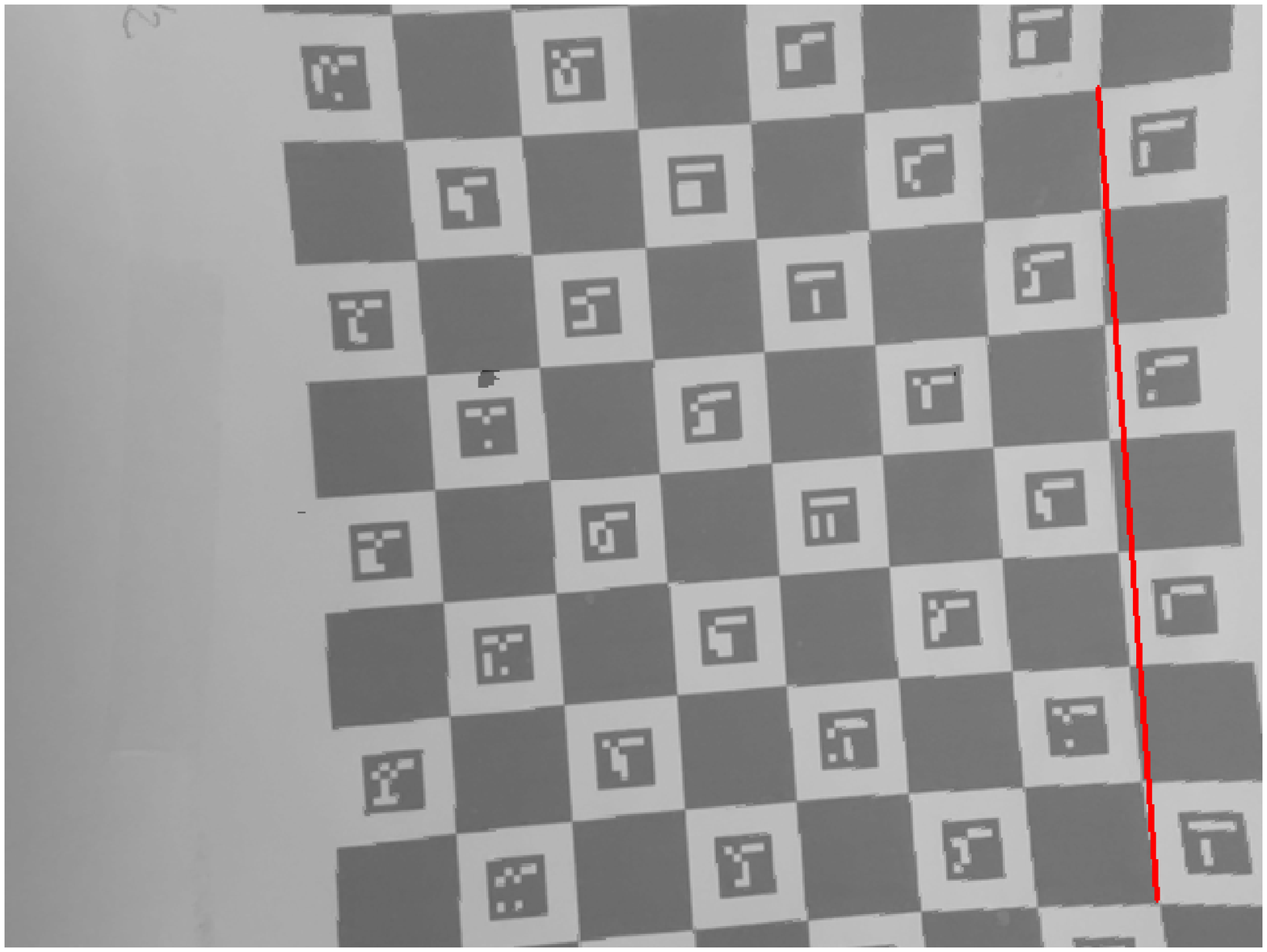}
    \caption{Applying \textit{LensDistortionLookupTable}}
    \label{fig:distortion_comparison_b}
 \end{subfigure}
  \begin{subfigure}[t]{0.3\textwidth}
    \includegraphics[width=1.0\textwidth]{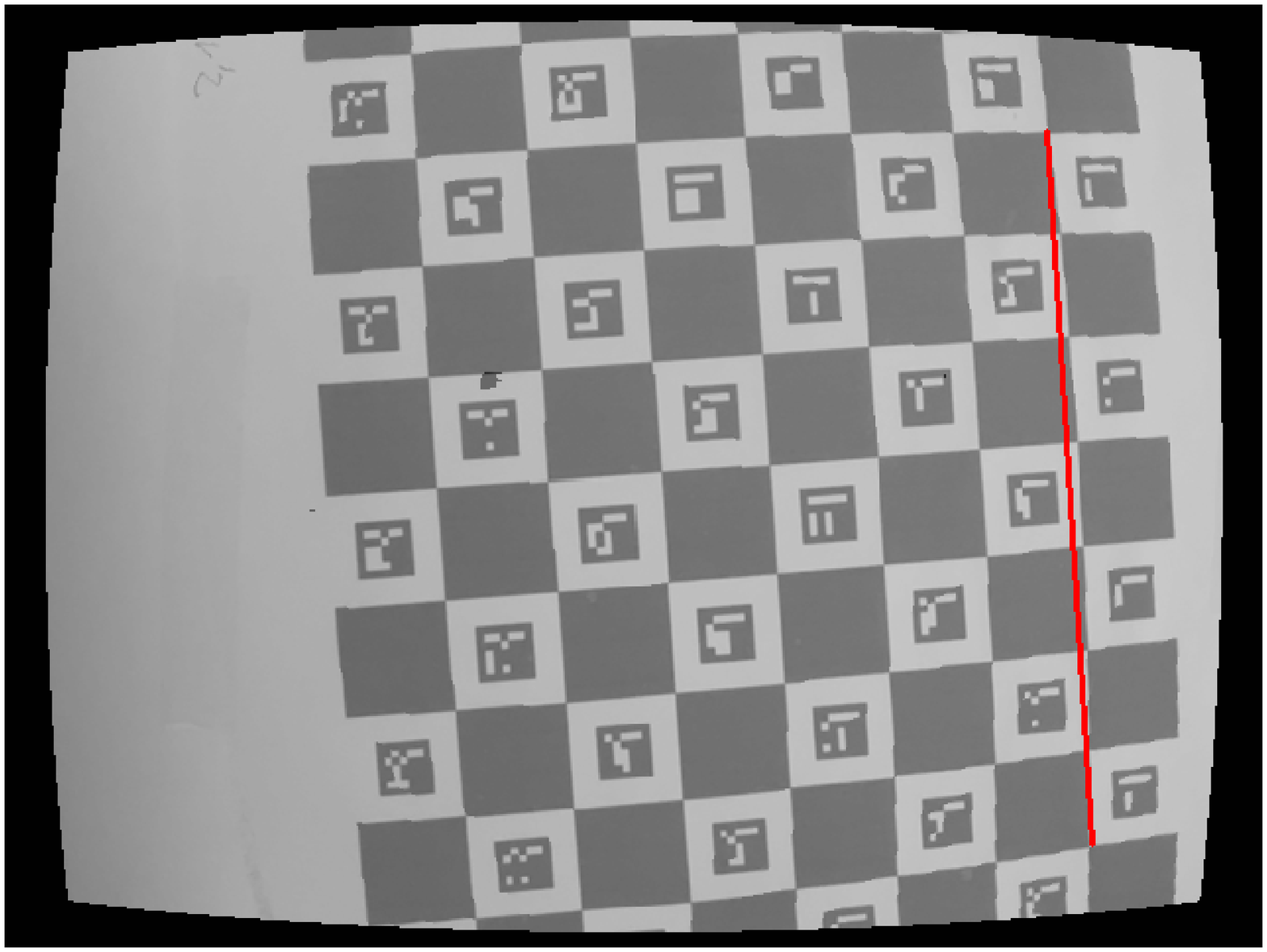}
    \caption{Applying \textit{LensDistortionInverseLookupTable}}
    \label{fig:distortion_comparison_c}
 \end{subfigure}
\caption{Example for an iPad 12.9'' Pro 5gen with a ultra wide angle front camera. To test if the image is already distortion corrected, we apply the forward distortion lookup table (b) and inverse distortion lookup table (c) to the original image (a) acquired from \textit{AVCaptureSession}. In the first case, the correction is applied twice, in the inverse case, we get the original distortion. This indicates, that all images are already distortion corrected.}
\label{fig:distortion_comparison}
\end{figure*}

In order to check if the depth maps or RGBD images need to be undistorted before usage, we can thus check the distortion of the RGB images. Figure \ref{fig:distortion_comparison} depicts a checkerboard recorded with an iPad 12.9'' Pro 5th generation. This iPad comes with an ultra wide lens and straight lines should be visibly distorted towards the image borders. Figure \ref{fig:distortion_comparison_a} shows the original image coming from a \textit{AVCaptureSession}. The image does not contain any visible distortion and straight lines appear to be perfectly straight. To evaluate if the provided distortion lookup tables have any effect we apply the forward (\textit{lensDistortionLookupTable}, figure \ref{fig:distortion_comparison_b}) and the inverse lookup table (\textit{inverseLensDistortionLookupTable}, figure \ref{fig:distortion_comparison_c}) to the original image as retrieved from \textit{ARKit} or \textit{AVFoundation} \footnote{\url{https://github.com/xybp888/iOS-SDKs/blob/master/iPhoneOS13.0.sdk/System/Library/Frameworks/AVFoundation.framework/Headers/AVCameraCalibrationData.h\#L118}}. Clearly the inverse lookup table maps the image back to an image containing the original distortion. The forward map seems to apply the distortion correction a second time. We repeated this experiment for all devices in table \ref{table:investigated_devices}. For some the initial lens distortion was so small that almost no effect was visible. 

\begin{displayquote}
\textbf{From our experiments, we conclude that all images recorded with the TrueDepth module from a AVCaptureSession or ARSession are already rectified and undistorted and do not need additional distortion correction.}
\end{displayquote}

This, however, is in contrast to another statement from the documentation\footnote{\url{https://developer.apple.com/documentation/avfoundation/avdepthdata/}}:

\begin{displayquote}
Because a depth data map is nonrectilinear, you can use an \textit{AVDepthData} map as a proxy for depth when rendering effects to its accompanying image, but not to correlate points in 3D space. To use depth data for computer vision tasks, use the data in the cameraCalibrationData property to rectify the depth data.
\end{displayquote}

We assume, that this might be valid for the disparity data recorded with other sensors like the TripleCamera, however, the verification is not in the scope of this study. For all tested RGB-D TrueDepth image pairs we could not identify any visible distortion, and hence assume that the data is already corrected.

\begin{table*}[t]
\centering
\begin{tabular}{|l|l|l|l|}
\hline
Device               & Factory $f_x$ {[}px{]} & Charuco $f_x$ {[}px{]} & Difference {[}\%{]} \\ \hline
iPhone 11 Pro V1     & 435.14               &  431.24          &  0.90      \\ \hline
iPhone 11 Pro V2     & 436.49               &  432.93          &  0.81      \\ \hline
iPhone 12            & 434.93               &  427.39          & 1.73       \\ \hline
iPhone 12 Pro        & 434.76               &  427.79          &  1.60      \\ \hline
iPhone 13            & 433.73               &  421.73          &  2.77      \\ \hline
                     &                      &                       &       \\ \hline
iPad 11'' 2gen       & 596.75               &  602.56          &  0.97      \\ \hline
\textbf{iPad 11'' 3gen}       & 571.28               &  532.60          &  \textbf{6.77}      \\ \hline
iPad 12.9'' Pro 4gen V1 & 597.76               &  591.49          &  1.10     \\ \hline
iPad 12.9'' Pro 4gen V2 & 595.33               &  602.51          &  1.20     \\ \hline
\textbf{iPad 12.9'' Pro 5gen} &  565.85              & 531.97           & \textbf{5.99}       \\ \hline
\end{tabular}
\caption{This table depicts a comparison of the factory calibrated focal length vs. a Charuco based calibration. The calibration was done with: {cv2.aruco.calibrateCameraCharuco()} and the following flags: CALIB\_FIX\_ASPECT\_RATIO, CALIB\_FIX\_K1, CALIB\_FIX\_K2, CALIB\_FIX\_K3, CALIB\_FIX\_TANGENT\_DIST, CALIB\_USE\_INTRINSIC\_GUESS, CALIB\_FIX\_PRINCIPAL\_POINT. Especially iPad 11'' 3gen and iPad 12.9'' Pro 5gen have quite large deviations in the range of 6-7\%. Interestingly this corresponds to the ratio between the unscaled depth intrinsics between AVSession and ARKit session (see table \ref{table:params_arsession} and table \ref{table:params_avsession}) which is ~7.5\%.}
\label{tab:cam_calib}
\end{table*}
%%%%%%%%%%%%%%%%%%%%%%%%%%%%%%%%%%%%%%%  
% List of device parameters
%%%%%%%%%%%%%%%%%%%%%%%%%%%%%%%%%%%%%%%  
%\include{differences_between_av_ar}

%%%%%%%%%%%%%%%%%%%%%%%%%%%%%%%%%%%%%%%  
% Solutions
%%%%%%%%%%%%%%%%%%%%%%%%%%%%%%%%%%%%%%% 
\section{Mitigating the Issues}
So far, we have identified two issues with iPad devices:
\begin{enumerate}
    \item \textbf{Misaligned depth maps in ARKit sessions} for iPad 11'' 2gen and iPad 12.9'' 4gen devices.
    \item \textbf{Wrong factory intrinsics} are returned from both APIs for iPad 11'' 3gen and iPad 12.9'' 5gen devices.
\end{enumerate}
In this section, we discuss two ways to fix the respective issue.

\subsection{Zooming the Depth Maps}
In order to correct the wrong depth to RGB mapping in the case of iPad 11'' 2gen and iPad 12'' 4gen models, we can calculate the difference in intrinsic reference dimensions (IRD) between AVSession and ARKit session. In all cases, this gives a factor of about 5.2\%: 
\begin{eqnarray}
zoom_x &=& \frac{IRD^{AV}_{width}}{IRD^{ARKit}_{width}} \approx 0.95\\
zoom_y &=& \frac{IRD^{AV}_{height}}{IRD^{ARKit}_{height}} \approx 0.95
\end{eqnarray} 

The Python code for re-mapping the depth map correctly to the RGB image is given in the following listing. Here we assume, that there is no aspect ratio and the zoom in width and height dimension is the same. In addition it is important to zoom to the principal point.
\begin{lstlisting}[language=python]
def zoom_in(depth_image, width, height, cx, cy, 
            ird_width_av = 3088, 
            ird_height_arkit = 3248):
    zoom_factor = ird_width_av / ird_height_arkit

    new_width = (zoom_factor * width)/2
    new_height = (zoom_factor * height)/2

    upper_left = [cx - new_width, cy - new_height]
    upper_right = [cx + new_width, cy - new_height]
    lower_right = [cx + new_width, cy + new_height]
    lower_left = [cx - new_width, cy + new_height]

    Pold = np.array([[0,0], [width,0], [width, height], [0,height]], dtype=np.float32)
    Pnew = np.array([upper_left, upper_right, lower_right, lower_left], dtype=np.float32)

    M = cv2.getPerspectiveTransform(Pold, Pnew)
    depth_zoomed = cv2.warpPerspective(depth_image, M, (depth.shape[1], depth.shape[0]))
    return depth_zoomed
\end{lstlisting}\label{code:zoom_py}

\subsection{Correcting the Focal Length}
In the case of iPad 12.9'' 5gen and iPad 11'' 3gen models the provided focal length is not correct. We found, that the focal length that \textbf{AVSession} returns is wrong by about 7\% (see \ref{tab:cam_calib}). However, if we compare the focal length between AVSession and ARKit session, we can again observe difference of about 7.5\%. But it seems that instead of a correction in the right direction (then we could maybe assume ARKit data is right) it seems that the focal length is again wrong by 7.5\%. \textbf{This doubles the effect when un-projected ARKit depth data is used for measurements or other computer vision applications.}.

To get the correct intrinsics for the depth camera, the following equation can be used to correct the meta data from AVSession:
\begin{equation}
    f^{corrected}_{av} = f^{unscaled\_depth}_{av} \frac{f^{unscaled\_depth}_{av}}{f^{depth}_{ar}} = \frac{(f^{unscaled\_depth}_{av})^2}{f^{depth}_{ar}}
\end{equation}
To then get the corrected focal length in VGA resolution (to use it with the depth images), we have to downscale the focal length:
\begin{equation}
    f^{scaled}_{av} = f^{corrected}_{av} \frac{640}{IRD^{av}_w}
\end{equation}

For example, to get the correctly scaled focal length for the iPad 12.9'' 5gen model in VGA resolution:
\begin{equation}
    \frac{1781.78^2px}{1916.17px} * \frac{640px}{2016px} = 525.97px
\end{equation}
This agrees within tolerance (~1\%) to the Charuco calibrated focal length for this iPad of about 531.97 pixels (depicted in table \ref{tab:cam_calib}). In the case that the data is recorded in ARKit mode the correction factor doubles. So we could use the following equation:

\begin{equation}
    f^{corrected}_{ar} = f^{unscaled\_depth}_{ar} (1+2*(1-\frac{f^{unscaled\_depth}_{ar}}{f^{depth}_{av}})) 
\end{equation}

and similarily the scaled focal length for VGA resolution:
\begin{equation}
    f^{scaled}_{ar} = f^{corrected}_{ar} \frac{640px}{IRD^{ar}_w}
\end{equation}
For example, to get the correctly scaled focal length for the iPad 12.9'' 5gen model in VGA resolution for data recorded in an ARKit session:
\begin{equation}
    1916.17px\times(1+2*(1-\frac{1916.17px}{1781.78px})) * \frac{640px}{2880px} = 361.58px
\end{equation}

The different focal length for AVSession (525.97) and for ARKit (361.58) also correspond well to the observed large field-of-view difference for iPad12.9'' 5gen and iPad 11'' 3gen as depicted in figure \ref{fig:overlay_iphone12_5gen_arkit}.
\label{sec:solutions}

%%%%%%%%%%%%%%%%%%%%%%%%%%%%%%%%%%%%%%%  
% Community Links
%%%%%%%%%%%%%%%%%%%%%%%%%%%%%%%%%%%%%%%  
%\input{community_links.tex}

\section{Discussion}
In this paper, we investigated the reliability of image and meta data of the TrueDepth camera across different iPad and iPhone devices and two different APIs. We showed that all iPad's have problems with the current APIs and that the camera data can not be reliably used for computer vision or measurement tasks without post-processing corrections. In addition, we presented two simple methods to correct the two different issues that we identified. Overall, we made several observations that might be relevant for developers: 
\begin{itemize}
\item \textbf{We do not expect this to be a hardware or factory calibration issue. It seems, that the correct data lies in the meta data of AVSession and ARKit sessions. So it is very likely a software bug.}
\item Aspect ratio was one in all tested devices and skew was always zero.
\item There was no visual distortion left, i.e. images and depth maps were already undistorted.
\item We presented two different ways to access the data using AVSession or ARKit. In ARKit sessions the data is mirrored around the up-axis.
\item The depth to RGB alignment is correct when data is recorded in AVSession. In most cases it is correct also in ARKit mode expect for iPad11'' 2gen and iPad12.9'' 4gen devices.
\item The alignment issue can be fixed by zooming the depth image with a factor computed from the different intrinsic reference dimensions between AVSession and ARKit session.
\item At the moment the TrueDepth camera of iPads should only be used with great care for computer vision or measuring tasks, due to the issues revealed in the paper.
\item Be aware that factory calibrated values seem to change from update to update. E.g. iPhone 11 Pro focal length changed by 0.5 pixel after updating to iOS15, and iPad 12.9'' intrinsics changed even more.
\item All iPhones have very well calibrated camera intrinsics and do not suffer from the two issues we observed for iPad devices.
\end{itemize}

\section{Bug Report}

We are aware of one forum post in Apple's developer forum from over a year ago \\ (\url{https://developer.apple.com/forums/thread/660388}) that describes one of the issues discussed in this report. \textbf{In addition to publishing this report, we opened an issue with Apple's feedback assistant (\url{https://feedbackassistant.apple.com/}) on Jan 17, 2022 (ticket id: FB9848651). Furthermore, we opened a TSI (technical support incident) on Jan 26, 2022. We will update this report as soon as the bug is fixed.}.

%\section*{Acknowledgements}

\include{table_tested_devices}
\begin{table}[]
\resizebox{\textwidth}{!}{%
\begin{tabular}{|l|l|l|l|l|l|l|l|}
\hline
Device               & \begin{tabular}[c]{@{}l@{}}Lens \\ distortion \\ center LDC \\{[}px{]}\end{tabular} & \begin{tabular}[c]{@{}l@{}}Intrinsics \\ reference \\ dimensions \\ {[}px{]}\end{tabular} & \begin{tabular}[c]{@{}l@{}}Depth\\ intrinsics\\ unscaled \\ {[}px{]}\end{tabular}                 & \begin{tabular}[c]{@{}l@{}}Color\\ intrinsics \\{[}px{]}\end{tabular} & \begin{tabular}[c]{@{}l@{}}LDC \\ equals \\ PP\end{tabular} & \begin{tabular}[c]{@{}l@{}}Depth \\ intrinsics \\ difference \\ to ARKit \end{tabular} & \begin{tabular}[c]{@{}l@{}}Intrinsics  \\ ref. dim. \\ difference \\ to ARKit \end{tabular} \\ \hline

\begin{tabular}[c]{@{}l@{}}iPad 11'' 3gen \\  V1\end{tabular} & \begin{tabular}[c]{@{}l@{}}x: 1011.80 \\ y: 754.60\end{tabular}  & \begin{tabular}[c]{@{}l@{}}W: 2016 \\ H: 1512\end{tabular}  & \begin{tabular}[c]{@{}l@{}}fx: 1791.13\\ fy: 1791.13\\ cx: 1012.10\\ cy: 754.53\end{tabular}  & \begin{tabular}[c]{@{}l@{}}fx: 568.61\\ fy: 568.61\\ cx: 321.23\\ cy: 239.47\end{tabular}  & \xmark & 7.5\% &42.0\% \\ \hline

\begin{tabular}[c]{@{}l@{}}iPad 11'' 3gen \\ V2\end{tabular} & \begin{tabular}[c]{@{}l@{}}x: 1013.80 \\ y: 757.30\end{tabular} &  \begin{tabular}[c]{@{}l@{}}W: 2016 \\ H: 1512\end{tabular} & \begin{tabular}[c]{@{}l@{}}fx: 1784.44\\ fy: 1784.44\\ cx: 1014.24\\ cy: 757.43\end{tabular}  & \begin{tabular}[c]{@{}l@{}}fx: 566.49\\ fy: 566.49\\ cx: 321.91\\ cy: 240.39\end{tabular}  & \xmark  & 7.5\% &42.0\%   \\ \hline

\begin{tabular}[c]{@{}l@{}}iPad 12.9'' 5gen \\ \textbf{iOS14} \end{tabular}  & \begin{tabular}[c]{@{}l@{}}x: 1009.73 \\ y: 759.42\end{tabular} & \begin{tabular}[c]{@{}l@{}}W: 2016 \\ H: 1512\end{tabular} & \begin{tabular}[c]{@{}l@{}}fx: \textbf{1781.78}\\ fy: \textbf{1781.78}\\ cx: 1009.89\\ cy: 759.69\end{tabular} & \begin{tabular}[c]{@{}l@{}}fx: \textbf{565.64}\\ fy: \textbf{565.64}\\ cx: 320.53\\ cy: 242.10\end{tabular} & \xmark & 7.5\% & 42.0\% \\ \hline

\begin{tabular}[c]{@{}l@{}}iPad 12.9'' 5gen \\ \textbf{After Update to iOS15}\end{tabular}  & \begin{tabular}[c]{@{}l@{}}x: 1009.73 \\ y: 759.42\end{tabular} & \begin{tabular}[c]{@{}l@{}}W: 2016 \\ H: 1512\end{tabular} & \begin{tabular}[c]{@{}l@{}}fx: \textbf{1780.13}\\ fy: \textbf{1780.13}\\ cx: 1009.89\\ cy: 759.69\end{tabular} & \begin{tabular}[c]{@{}l@{}}fx: \textbf{565.12}\\ fy: \textbf{565.12}\\ cx: 320.53\\ cy: 242.10\end{tabular} & \xmark & 7.5\% & 42.0\% \\ \hline\hline

\begin{tabular}[c]{@{}l@{}}iPad 11'' 2gen  \\V1\end{tabular} & \begin{tabular}[c]{@{}l@{}}x: 1543.98\\ y: 1147.17\end{tabular}  & \begin{tabular}[c]{@{}l@{}}W: 3088\\ H: 2316\end{tabular}  & \begin{tabular}[c]{@{}l@{}}fx: 2878.39\\ fy: 2878.39\\ cx: 1543.98\\ cy: 1147.17\end{tabular} & \begin{tabular}[c]{@{}l@{}}fx: 596.55\\ fy: 596.55\\ cx: 319.60\\ cy: 237.36\end{tabular} & \cmark & 0.0\% & 5.2\%   \\ \hline

\begin{tabular}[c]{@{}l@{}}iPad 11''  2gen \\V2\end{tabular}  & \begin{tabular}[c]{@{}l@{}}x: 1541.02\\ y: 1155.53\end{tabular}  & \begin{tabular}[c]{@{}l@{}}W: 3088\\ H: 2316\end{tabular}  & \begin{tabular}[c]{@{}l@{}}fx: 2882.66\\ fy: 2882.66\\ cx: 1541.02\\ cy: 1155.53\end{tabular} & \begin{tabular}[c]{@{}l@{}}fx: 597.44\\ fy: 597.44\\ cx: 318.98\\ cy: 239.09\end{tabular} & \cmark & 0.0\% & 5.2\%  \\ \hline

\begin{tabular}[c]{@{}l@{}}iPad 12.9'' 4gen\\  V1\end{tabular} & \begin{tabular}[c]{@{}l@{}}x: 1538.43\\ y: 1151.34\end{tabular}  & \begin{tabular}[c]{@{}l@{}}W: 3088\\ H: 2316\end{tabular}  & \begin{tabular}[c]{@{}l@{}}fx: 2885.64\\ fy: 2885.64\\ cx: 1538.43\\ cy: 1151.34\end{tabular} & \begin{tabular}[c]{@{}l@{}}fx: 598.06\\ fy: 598.06\\ cx: 318.45\\ cy: 238.22\end{tabular} & \cmark & 0.0\% & 5.2\% \\ \hline

\begin{tabular}[c]{@{}l@{}}iPad 12.9'' 4gen\\  V2\end{tabular} & \begin{tabular}[c]{@{}l@{}}x: 1535.93\\ y: 1157.79\end{tabular}  & \begin{tabular}[c]{@{}l@{}}W: 3088\\ H: 2316\end{tabular}  & \begin{tabular}[c]{@{}l@{}}fx: 2872.52\\ fy: 2872.52\\ cx: 1535.93\\ cy: 1157.79\end{tabular} & \begin{tabular}[c]{@{}l@{}}fx: 595.33\\ fy: 595.33\\ cx: 317.93\\ cy: 239.56\end{tabular} & \cmark & 0.0\% & 5.2\% \\ \hline\hline

\begin{tabular}[c]{@{}l@{}}iPhone 11 Pro \\ \textbf{iOS14}\end{tabular}        & \begin{tabular}[c]{@{}l@{}}x: 2026.13\\ y: 1508.71\end{tabular} & \begin{tabular}[c]{@{}l@{}}W: 4032\\ H: 3024\end{tabular} & \begin{tabular}[c]{@{}l@{}}fx: \textbf{2751.18}\\ fy: \textbf{2751.18}\\ cx: 2026.13\\ cy: 1508.71\end{tabular} & \begin{tabular}[c]{@{}l@{}}fx: \textbf{437.29}\\ fy: \textbf{437.29}\\ cx: 321.18\\ cy: 239.05\end{tabular} & \cmark & 0.00\% & 0.0\% \\ \hline

\begin{tabular}[c]{@{}l@{}}iPhone 11 Pro \\ \textbf{After Update to iOS15}\end{tabular}& \begin{tabular}[c]{@{}l@{}}x: 2026.13\\ y: 1508.71\end{tabular} & \begin{tabular}[c]{@{}l@{}}W: 4032\\ H: 3024\end{tabular} & \begin{tabular}[c]{@{}l@{}}fx: \textbf{2750.68}\\ fy: \textbf{2750.68}\\ cx: 2026.13\\ cy: 1508.71\end{tabular} & \begin{tabular}[c]{@{}l@{}}fx: \textbf{436.61}\\ fy: \textbf{436.61}\\ cx: 321.18\\ cy: 239.05\end{tabular} & \cmark & 0.00\% & 0.0\% \\ \hline

\end{tabular} }
\caption{This table subsumes the TrueDepth camera parameters for \textbf{\textit{AVSession}}. V1 and V2 correspond to different devices from the same generation. An interesting observation is, that the focal length of the iPhone 11 Pro and iPad12.9'' 5gen changed after updating from iOS14 to iOS15.}
\label{table:params_avsession}
\end{table}
\begin{table}[]
\resizebox{\textwidth}{!}{%
\begin{tabular}{|l|l|l|l|l|l|}
\hline
Device               & \begin{tabular}[c]{@{}l@{}}Lens \\ distortion \\ center LDC \\{[}px{]}\end{tabular} & \begin{tabular}[c]{@{}l@{}}Intrinsics \\ reference \\ dimensions \\{[}px{]}\end{tabular} & \begin{tabular}[c]{@{}l@{}}Depth\\ intrinsics\\ unscaled \\{[}px{]}\end{tabular}                 & \begin{tabular}[c]{@{}l@{}}Color\\ intrinsics {[}px{]}\end{tabular} & \begin{tabular}[c]{@{}l@{}}LDC \\ equals \\ PP\end{tabular} \\ \hline

\begin{tabular}[c]{@{}l@{}}iPad 11'' Pro 3gen \\ V1 \end{tabular} & \begin{tabular}[c]{@{}l@{}}x: 1428.30 \\ y: 1080.89\end{tabular}  & \begin{tabular}[c]{@{}l@{}}W: 2880 \\ H: 2160\end{tabular}  & \begin{tabular}[c]{@{}l@{}}fx: 1925.71\\ fy: 1925.71\\ cx: 1426.63\\ cy: 1080.95\end{tabular}  & \begin{tabular}[c]{@{}l@{}}fx: 962.86\\ fy: 962.86\\ cx: 713.06\\ cy: 539.72\end{tabular}  & \xmark  \\ \hline

\begin{tabular}[c]{@{}l@{}}iPad 11'' Pro 3gen \\ V2 \end{tabular}  & \begin{tabular}[c]{@{}l@{}}x: 1428.80 \\ y: 1078.69\end{tabular} &  \begin{tabular}[c]{@{}l@{}}W: 2880 \\ H: 2160\end{tabular} & \begin{tabular}[c]{@{}l@{}}fx: 1917.50\\ fy: 1917.50\\ cx: 1427.21\\ cy: 1078.42\end{tabular}  & \begin{tabular}[c]{@{}l@{}}fx: 958.75\\ fy: 958.75\\ cx: 713.35\\ cy: 538.46\end{tabular}  & \xmark     \\ \hline

\begin{tabular}[c]{@{}l@{}}iPad 12.9'' Pro 5gen \\ \textbf{iOS14} \end{tabular} & \begin{tabular}[c]{@{}l@{}}x: 1426.73\\ y: 1077.07\end{tabular} & \begin{tabular}[c]{@{}l@{}}W: 2880 \\ H: 2160\end{tabular} & \begin{tabular}[c]{@{}l@{}}fx: \textbf{1916.17}\\ fy: \textbf{1916.17}\\ cx: 1424.82\\ cy: 1076.56\end{tabular} & \begin{tabular}[c]{@{}l@{}}fx: \textbf{958.08}\\ fy: \textbf{958.08}\\ cx: 712.16\\ cy: 537.53\end{tabular} & \xmark \\ \hline

\begin{tabular}[c]{@{}l@{}}iPad 12.9'' Pro 5gen \\ \textbf{After update from iOS 14 to 15} \end{tabular} & \begin{tabular}[c]{@{}l@{}}x: 1426.73\\ y: 1077.07\end{tabular} & \begin{tabular}[c]{@{}l@{}}W: 2880 \\ H: 2160\end{tabular} & \begin{tabular}[c]{@{}l@{}}fx: \textbf{1914.39}\\ fy: \textbf{1914.39}\\ cx: 1424.82\\ cy: 1076.56\end{tabular} & \begin{tabular}[c]{@{}l@{}}fx: \textbf{957.19}\\ fy: \textbf{957.19}\\ cx: 712.16\\ cy: 537.53\end{tabular} & \xmark \\ \hline\hline

\begin{tabular}[c]{@{}l@{}}iPad 11'' 2gen \\ V1 \end{tabular}  & \begin{tabular}[c]{@{}l@{}}x: 1624.48\\ y: 1226.32\end{tabular}  & \begin{tabular}[c]{@{}l@{}}W: 3248\\ H: 2436\end{tabular}  & \begin{tabular}[c]{@{}l@{}}fx: 2878.39\\ fy: 2878.39\\ cx: 1624.48\\ cy: 1226.32\end{tabular} & \begin{tabular}[c]{@{}l@{}}fx: 1276.13\\ fy: 1276.13\\ cx: 719.93\\ cy: 542.96\end{tabular} & \cmark \\ \hline

\begin{tabular}[c]{@{}l@{}}iPad 11'' 2gen \\ V2 \end{tabular} & \begin{tabular}[c]{@{}l@{}}x:1621.52\\ y: 1217.96\end{tabular}  & \begin{tabular}[c]{@{}l@{}}W: 3248\\ H: 2436\end{tabular}  & \begin{tabular}[c]{@{}l@{}}fx: 2882.66\\ fy: 2882.66\\ cx: 1621.52\\ cy: 1217.96\end{tabular} & \begin{tabular}[c]{@{}l@{}}fx: 1278.02\\ fy: 1278.02\\ cx: 718.62\\ cy: 539.26\end{tabular} &  \cmark \\ \hline

\begin{tabular}[c]{@{}l@{}}iPad 12.9'' 4gen \\ V1 \end{tabular} & \begin{tabular}[c]{@{}l@{}}x: 1618.93\\ y: 1222.15\end{tabular}  & \begin{tabular}[c]{@{}l@{}}W: 3248\\ H: 2436\end{tabular}  & \begin{tabular}[c]{@{}l@{}}fx: 2885.64\\ fy: 2885.64\\ cx: 1618.93\\ cy: 1222.15\end{tabular} & \begin{tabular}[c]{@{}l@{}}fx: 1279.35\\ fy: 1279.35\\ cx: 717.47\\ cy: 541.12\end{tabular} &  \cmark \\ \hline\hline

\begin{tabular}[c]{@{}l@{}}iPad 12.9'' 4gen \\ V2 \end{tabular} & \begin{tabular}[c]{@{}l@{}}x: 1616.43\\ y: 1215.70\end{tabular}  & \begin{tabular}[c]{@{}l@{}}W: 3248\\ H: 2436\end{tabular}  & \begin{tabular}[c]{@{}l@{}}fx: 2872.51\\ fy: 2872.51\\ cx: 1616.43\\ cy: 1215.70\end{tabular} & \begin{tabular}[c]{@{}l@{}}fx: 1273.52\\ fy: 1273.52\\ cx: 716.36\\ cy: 538.26\end{tabular} &  \cmark \\ \hline\hline

\begin{tabular}[c]{@{}l@{}}iPhone 11 Pro \\ \textbf{iOS 14} \end{tabular}        & \begin{tabular}[c]{@{}l@{}}x: 2026.63\\ y: 1514.78\end{tabular} & \begin{tabular}[c]{@{}l@{}}W: 4032\\ H: 3024\end{tabular} & \begin{tabular}[c]{@{}l@{}}fx: \textbf{2751.18}\\ fy: \textbf{2751.18}\\ cx: 2026.63\\ cy: 1514.78\end{tabular} & \begin{tabular}[c]{@{}l@{}}fx:  \textbf{982.56}\\ fy: \textbf{982.56}\\ cx: 723.47\\ cy: 540.31\end{tabular} & \cmark \\ \hline

\begin{tabular}[c]{@{}l@{}}iPhone 11 Pro \\ \textbf{After update from iOS 14 to 15} \end{tabular}  & \begin{tabular}[c]{@{}l@{}}x: 2026.63\\ y: 1514.78\end{tabular} & \begin{tabular}[c]{@{}l@{}}W: 4032\\ H: 3024\end{tabular} & \begin{tabular}[c]{@{}l@{}}fx: \textbf{2750.68}\\ fy: \textbf{2750.68}\\ cx: 2026.63\\ cy: 1514.78\end{tabular} & \begin{tabular}[c]{@{}l@{}}fx:  \textbf{982.38}\\ fy: \textbf{982.38}\\ cx: 723.47\\ cy: 540.31\end{tabular} & \cmark \\ \hline
\end{tabular}
}
\caption{This table subsumes the TrueDepth camera parameters for \textbf{\textit{ARKitSession}}. V1 and V2 correspond to different devices from the same generation. An interesting observation is, that the focal length of the iPhone 11 Pro and the iPad12.9'' 5gen changed after updating from iOS14 to iOS15.}
\label{table:params_arsession}
\end{table}

\bibliography{bibliography}
\end{document}